\documentclass{article}

\usepackage[accepted]{icml2015}

\newcommand*{\getTitle}{Variational Inference with Hamiltonian Monte Carlo}
\newcommand*{\getShortTitle}{Variational Inference with Hamiltonian Monte Carlo}

\newcommand*{\getUniversityTUM}{Technische Universit\"at M\"unchen}

\newcommand{\E}{\mathbb{E}}

\newcommand{\Cov}{\mathrm{Cov}}

\definecolor{m0}{rgb}{0.368417, 0.506779, 0.709798} 
\definecolor{m1}{rgb}{0.880722, 0.611041, 0.142051} 
\definecolor{m2}{rgb}{0.560181, 0.691569, 0.194885}
\definecolor{m3}{rgb}{0.922526, 0.385626, 0.209179}
\definecolor{m4}{rgb}{0.528488, 0.470624, 0.701351}
\definecolor{m5}{rgb}{0.772079, 0.431554, 0.102387}
\definecolor{m6}{rgb}{0.363898, 0.618501, 0.782349}
\definecolor{m7}{rgb}{1, 0.75, 0}
\definecolor{m8}{rgb}{0.647624, 0.37816, 0.614037}
\definecolor{m9}{rgb}{0.571589, 0.586483, 0.}
\definecolor{m10}{rgb}{0.915, 0.3325, 0.2125}
\definecolor{m11}{rgb}{0.400822, 0.522007, 0.85}
\definecolor{m12}{rgb}{0.972829, 0.621644, 0.073362}
\definecolor{m13}{rgb}{0.736783, 0.358, 0.503027}
\definecolor{m14}{rgb}{0.280264, 0.715, 0.429209}

\newcommand{\tn}{\tabularnewline}

\PassOptionsToPackage{table,svgnames,dvipsnames}{xcolor}

\usepackage[%
  backend=bibtex,
  url=false,
  style=authoryear,
  maxnames=2,
  minnames=1,
  maxbibnames=99,
  firstinits,
  hyperref=true,
  doi=false,
  isbn=false,
  url=false,
  uniquename=init]{biblatex} 
\usepackage{graphicx}
\usepackage{tikz}
\usetikzlibrary{arrows}
\usetikzlibrary{shapes}
\usepackage{tikzscale}
\usepackage{array}
\usepackage{booktabs}
\usepackage{hyperref} 
\usepackage[textwidth=1.5cm, textsize=tiny]{todonotes}
\usepackage{amsmath}
\usepackage{mathtools}
\usepackage{amssymb}
\usepackage{amsthm}
\usepackage{algorithm,algpseudocode}
\usepackage[titletoc,toc,title,page]{appendix}
\usepackage[symbol]{footmisc}
\DeclareFixedFont{\ttb}{T1}{txtt}{bx}{n}{9} 
\DeclareFixedFont{\ttm}{T1}{txtt}{m}{n}{9}  

\bibliography{bibliography/library}
\icmltitlerunning{\getShortTitle}

\begin{document} 

\twocolumn[
\icmltitle{\getTitle{}}


\icmlauthor{Christopher Wolf}{christopher.wolf@tum.de}
\icmladdress{}
\icmlauthor{Maximilian Karl\footnotemark[1] \& Patrick van der Smagt\footnotemark[1]}{karlma@in.tum.de}
\icmladdress{Chair of Robotics and Embedded Systems, Department of Informatics,\\\getUniversityTUM, Germany}

\icmlkeywords{Variational Inference, Hamiltonian Monte Carlo, density estimation, Variational Auto Encoder}
\vskip 0.3in
]
\footnotetext[1]{Patrick van der Smagt and Maximilian Karl are also affiliated with fortiss, An-Institut der Technischen Universit\"at M\"unchen, Germany}
\begin{abstract} 
Variational inference lies at the core of many state-of-the-art algorithms. To improve the approximation of the posterior beyond parametric families, it was proposed to include MCMC steps into the variational lower bound. In this work we explore this idea using steps of the Hamiltonian Monte Carlo (HMC) algorithm, an efficient MCMC method. In particular, we incorporate the acceptance step of the HMC algorithm, guaranteeing asymptotic convergence to the true posterior. Additionally, we introduce some extensions to the HMC algorithm geared towards faster convergence. The theoretical advantages of these modifications are reflected by performance improvements in our experimental results.
\end{abstract}

\setcounter{footnote}{0}
\renewcommand*{\thefootnote}{\arabic{footnote}}
\section{Introduction}

In modern data analysis probabilistic graphical models have emerged as a powerful and intuitive tool to capture and reveal hidden structures present in the data. Training and interpreting these models requires inferring the hidden variables of the observed data under the model. In many state-of-the-art graphical model approaches this key task is performed based on variational inference, a method converting complex inference problems into high-dimensional optimization problems \parencite{Jordan1999}. For instance \textcite{Hoffman2013} follow this approach for large scale text-to-topic models and \textcite{Gregor2015, Rezende2014, Kingma2014} apply it to the generation of images. 

Variational inference approximates the intractable true posterior distribution by the best-fitting candidate from a fixed family of distributions. While this makes the approximation procedure very fast, the restriction to a usually quite limited family of distributions means, that often the true posterior is only poorly approximated. This in turn hampers the training and final performance of the graphical model. Many suggestions for broader families of candidate distributions have been put forward allowing for more complicated approximations. A powerful framework, unifying several previous approaches, is the work by \textcite{Rezende2015} on normalizing flows. Here, arbitrarily complicated distributions are generated by applying a sequence of invertible mappings to a simple initial distribution. An interesting example for such a normalizing flow is the Hamiltonian variational inference method derived by \textcite{Salimans2014}, where steps of the Hamiltonian Monte Carlo (HMC) algorithm are used to transform the initial distribution. Since the HMC algorithm generates a Markov chain converging to the true posterior, this extension to variational inference is particularly appealing, because the generated family of distributions is guaranteed to contain the true posterior (provided enough steps are taken). However, \textcite{Salimans2014} left out the acceptance step of the HMC algorithm, so that convergence to the true posterior is no longer ensured and the true posterior need not be within the generated distribution family.

In this work we exploit the structure of the HMC algorithm to derive the variational lower bound for the case, where a distribution is transformed by steps of the full HMC algorithm including the acceptance step. By doing so, we regain the asymptotic guarantee of a perfect approximation. Additionally, we present two extensions to the HMC algorithm, which can be included in the approximation procedure and speed up the convergence to the true posterior. We begin by revising variational inference, MCMC methods and the work by \textcite{Salimans2014} on their combination (section~\ref{sec:VIandMCMC}) as well as the HMC algorithm (section~\ref{sec:HMC}). In section~\ref{sec:HMCVI} the aforementioned extensions to the variational lower bound are derived, before being applied in section~\ref{sec:Experiments}. In the final section~\ref{sec:ConclAndFuture} some ideas for further improvements are discussed.
\section{Variational inference and MCMC}
\label{sec:VIandMCMC}
\subsection{Variational inference}

In a probabilistic model $p(x, z)$ with missing or latent variables $z$ (possibly parameters in a Bayesian setting) the quantity of interest for inference problems is the marginal likelihood $p(x) = \int p(x, z) dz$. This integral is usually intractable and only a lower bound $\mathcal{L}$ to its value can be obtained using the variational principle:
\begin{equation}
\begin{split}
\log p(x) &\geq \log p(x) - D_{KL} \left( q_{\theta}(z|x) || p(z|x) \right) \\
			   &=  \E_{q_{\theta}(z|x)} \left[ \log p(x, z) - \log q_\theta(z|x) \right] \eqqcolon \mathcal{L}
\end{split}
\end{equation}
This requires the approximation of the true posterior $p(z|x)$, which is usually also intractable, by a parametrized density $q_{\theta}(z|x)$. By maximizing $\mathcal{L}$ with respect to the parameters $\theta$, the KL-divergence between the true and the approximate posterior is minimized and reaches its minimum, when the approximation equals the true posterior. In this case, $\log p(x) = \mathcal{L}$. From this derivation it is clear, that the success of this method, known as variational inference (VI), strongly depends on the approximation capacity of $q_\theta$.

\subsection{MCMC}
\label{sec:MCMC}


A widely used method to approximate intractable distributions is to repeatedly sample from them using Markov Chain Monte Carlo (MCMC) methods. To draw samples from an arbitrary target distribution with density $f_\textrm{target}(s)$ using MCMC, first a random state $s_0$ is drawn from some initial distribution $q_0(s)$. Then, a stochastic transition operator $s_{t} \sim q(s_t|s_{t-1})$ is applied repeatedly, producing a Markov chain $(s_t)_{t \in \mathbb{N}}$. By appropriate choice of the transition density $q$ a Markov chain can be constructed, which under minor regularity conditions has two key properties: Firstly its stationary distribution is the target distribution $f_\textrm{target}$ and secondly the chain converges to its stationary distribution \parencite{Roberts2004}. Therefore, by running such a chain for a sufficient number of steps, a sample from the target distribution can be obtained. However, the number of steps required is unknown a priori and may be very large.

The most common method for constructing such a Markov chain is the Metropolis-Hastings algorithm, where the transition is constructed in two steps: First a new proposed state $\tilde{s}_t$ is sampled from a proposal distribution $\tilde{q}(\tilde{s}_t|s_{t-1})$. In the second step, the acceptance step, this proposal is then accepted as the new state with probability 
\begin{equation} \label{eq:Metropolis-Hastings}
\begin{split}
p_{\textrm{accept}}&(s_{t-1}, \tilde{s}_t) \\
&= \min \left[ 1, \frac{f_\textrm{target}(\tilde{s}_t)}{f_\textrm{target}(s_{t-1})} \cdot \frac{\tilde{q}(s_{t-1}|\tilde{s}_t)}{\tilde{q}(\tilde{s}_t|s_{t-1})} \right],
\end{split}
\end{equation}
in which case we set $s_t = \tilde{s}_t$. Otherwise, the current state is kept, so $s_t = s_{t-1}$. It can be shown, that this indeed produces a Markov chain with the required properties \parencite{Roberts2004}. 

It is important to note that the target distribution density appears both in the enumerator and denominator, so we do not need the target distribution function to be normalized. This is essential for the use of MCMC with Bayesian inference, since Bayes's Theorem states $p(z|x) \propto p(x|z) \cdot p(z)$ with the usually intractable normalization factor $p(x)$.

\subsection{Combining variational inference and MCMC}
\label{sec:MCVI}
For sampling from the intractable posterior $p(z|x)$ via MCMC, we could choose the unobserved variable $z$ as state and the exact posterior $p(z|x)$ as target distribution. In contrast to the parametrized distribution $q_\theta(z|x)$ in VI, this gives us an asymptotically exact approximation of the posterior. However, it is also computationally expensive and does not offer an explicit objective function (which is e.g.\ needed for training the generative model $p(x, z)$).

To integrate the adaptiveness of MCMC into VI \textcite{Salimans2014} have proposed a powerful combination of these two methods, which they call Markov Chain Variational Inference (MCVI). The idea is to interpret the Markov chain obtained in MCMC as a variational approximation $q(z_0, \dots, z_T|x) = q_{0}(z_0|x) \cdot \prod_{t=1}^T q(z_t|z_{t-1}, x)$. Due to the additional variables $y = (z_0, \dots, z_{T-1})$ ($z_T$ corresponds to the output of standard VI), the lower bound must be modified:
\begin{equation}
\begin{split}
\log &p(x) \geq \mathcal{L} \\
	   &\geq \mathcal{L} - \E_{q(z_T|x)} \big[ D_{KL} \left( q(y |z_T, x) || r(y|z_T, x) \right) \big] \\
	   &=  \E_{q(y, z_T|x)} \big[ \log p(x, z_T) + \log r(y |z_T, x) \\
	   &\qquad\qquad\qquad  - \log q(y, z_{T} |x) \big] \\
	   &\eqqcolon \mathcal{L}_{\textrm{aux}},
\end{split}
\end{equation}
where $r(y|z_T, x)$ is an auxiliary distribution to be learnt as an approximation of the intractable $q(y |z_T, x)$. 

Due to the Markov chain structure of the \textit{forward} distribution $q(z_1, \dots, z_T|z_0, x) = \prod_{t=1}^T q(z_t|z_{t-1}, x)$, a natural choice for the auxiliary \textit{reverse} distribution is to mimic this structure, i.e.\ to assume $r(z_0, \dots, z_{T-1} |z_T, x) = \prod_{t=1}^T r(z_{t-1}|z_t, t, x)$. It is worth noting that conversely to the forward model, where the transitions should be independent of the step number (as in MCMC), the reverse model may use the step number to achieve a better fit. This allows the reverse model to capture the decreasing bias due to the initial distribution $q_0(z_0|x)$. In this case, the auxiliary lower bound can be rewritten as
\begin{equation} \label{eq:MCVIAuxLowerBound}
\begin{split}
\mathcal{L}_{\textrm{aux}} &= \E_{q(z_0, \dots, z_T|x)} \left[ \log p(x, z_T) - \log q(z_0|x) \right] \\
& \quad + \sum_{t=1}^T \E_{q(z_0, \dots, z_T|x)} \big[ \log r(z_{t-1}|z_t, t, x) \\
& \quad\qquad\qquad\qquad\qquad\; - \log q(z_t|z_{t-1}, x)  \big] 
\end{split}
\end{equation}

Provided that the random variables within the expectations are differentiable w.r.t.\ the parameters, an efficient Monte Carlo estimate of the gradient of the lower bound w.r.t.\ the parameters can be computed \parencite{Kingma2014, Rezende2014}. This gradient estimate can then be used to train the forward and the reverse model (and if applicable the generative model $p(x, z)$) using gradient-based stochastic optimization algorithms such as Adam \parencite{Kingma2015}.

\section{Hamiltonian Monte Carlo}
\label{sec:HMC}
A very popular MCMC method is the Hamiltonian Monte Carlo (or Hybrid Monte Carlo, HMC) algorithm \parencite{Duane1987}, since it is highly efficient and widely applicable. The idea behind this algorithm is to propose new points by simulating the dynamics of a particle on a potential energy landscape induced by the desired target distribution. This simulation is done using the Hamiltonian dynamics formulation, which results in several useful properties for the HMC algorithm. These can be further exploited by using HMC within the MCVI scheme. To understand these synergies, we will first review Hamiltonian dynamics and the HMC algorithm. For a more exhaustive review and discussion refer to \textcite{Neal2011}.

\subsection{Hamiltonian dynamics}

Hamiltonian dynamics (HD) is a reformulation of classical dynamics, where the state of the physical system is described by a pair $(q, p)$ of $d$-dimensional vectors, where $q$ is the \textit{position} vector and $p$ is the \textit{momentum} vector. The evolution of the system through time is then given by \textit{Hamilton's equations}:
\begin{equation} \label{eq:HamiltonsEquations}
\begin{split}
\frac{dq_i}{dt} &= \frac{\partial H}{\partial p_i} \\
\frac{dp_i}{dt} &= - \frac{\partial H}{\partial q_i},
\end{split}
\end{equation}
where $H(q, p, t)$ is the \textit{Hamiltonian} of the system (often its total energy).

\begin{figure*}
\centering
\includegraphics[width=2.05\columnwidth]{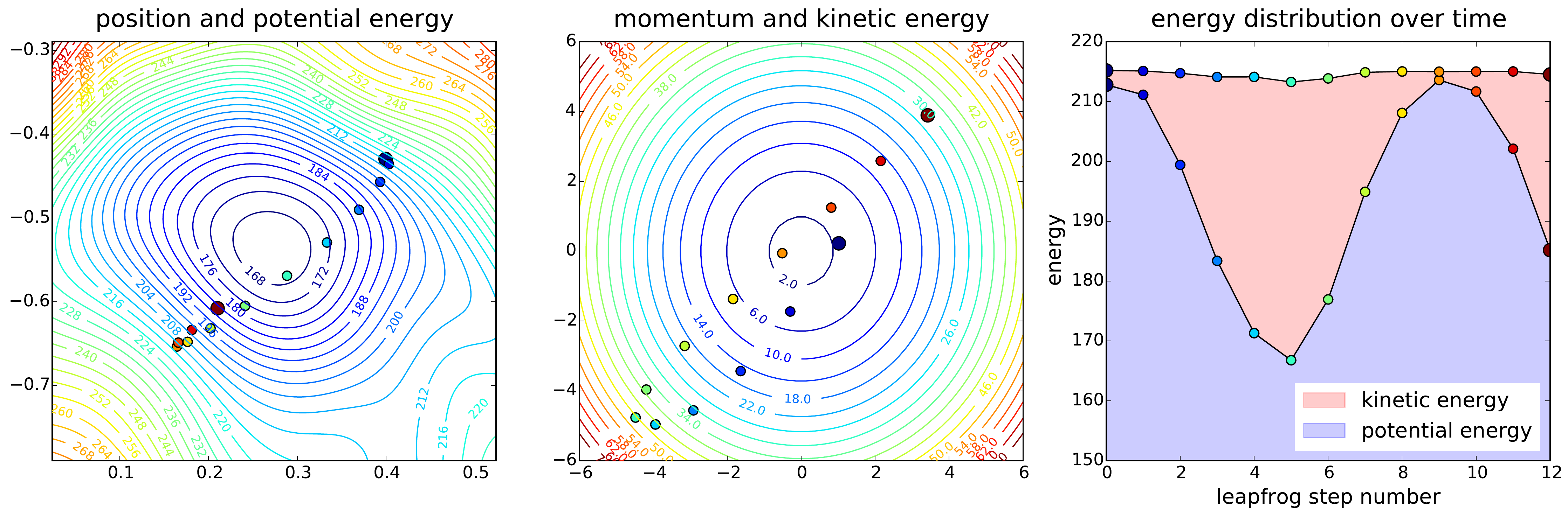}
\caption{Dynamics of a particle under HD computed using the leapfrog method. Each computed point along the discretized trajectory is indicated by a separate color ranging from dark blue (starting point) to dark red (final point). The left plot shows the position of the particle with the prescribed potential energy represented by the contour plot. The centre plot depicts the momentum of the particle with the kinetic energy at each point indicated by the contours. In the plot on the right the energy distribution of the particle over time is given, with the potential energy in blue and the kinetic energy in red. Due to the discretization the total energy is not exactly conserved.}
\label{fig:HMC_MOTION_1hmc_12lf}
\end{figure*}

For our application, we are interested in the motion of a frictionless particle governed by the \textit{potential energy} $U(q)$ and \textit{kinetic energy} $K(p)$. In this setting the Hamiltonian is just the total energy of the system, i.e.\ $H(q, p) = U(q) + K(p)$, which is independent of time due to conservation of energy. In two dimensions this can be visualized well as a frictionless particle sliding over a landscape of varying height (see figure~\ref{fig:HMC_MOTION_1hmc_12lf} for a numerically solved example).

In such a physical system the kinetic energy is then given by $K(p) = p^T M^{-1} p /2$, where $M$ is called the mass matrix and in a physical context usually is $m I$, a scalar multiple of the identity. Here, the scalar $m$ corresponds to the mass of the particle. With this kinetic energy we can retrieve Newton's equation of motion relating the acceleration $d^2q/dt^2$ to the force acting on a particle (given by $-\nabla U(q)$):
\begin{equation} \label{eq:NewtonsEquation}
\frac{d^2q}{dt^2} = M^{-1} \frac{dp}{dt} = - M^{-1} \frac{\partial H}{\partial q} = - M^{-1} \nabla U(q)
\end{equation}

The key advantage of HD over other formulations of classical dynamics is that analytic solutions to Hamilton's equations \eqref{eq:HamiltonsEquations} have three crucial properties \parencite{Neal2011}:
\begin{itemize}
\item Reversibility: The mapping $T_s$ from the state $(q(t), p(t))$ at some time point $t$ to the state at $t+s$ ($s > 0$) is one-to-one and hence reversible. Thus by running time backwards, i.e.\ negating both time derivatives in Hamilton's equations, we can uniquely determine previous states.
\item Volume preservation: $T_s$ conserves volume in $(q, p)$-space, so applying it to some region of a certain volume results in a region of the same volume.
\item Conservation of the Hamiltonian: The Hamiltonian $H(q, p)$ is invariant with time, so $dH/dt = 0$.
\end{itemize}

All three of these properties would be useful in the application of the HMC algorithm, but not all of them can be preserved in numerical solutions of~\eqref{eq:HamiltonsEquations}. The leapfrog method, which will be explained below, yields numerical solutions which maintain reversibility and volume preservation and furthermore approximately conserve the Hamiltonian (see figure~\ref{fig:HMC_MOTION_1hmc_12lf}). This approximate conservation of the Hamiltonian makes the leapfrog method a so-called \textit{symplectic integrator}.

Given the step size $\epsilon$ the leapfrog method performs the following discrete updates for $n \in \mathbb{N}_0$ starting from the initial state $(q^{(0)}, p^{(0)})$:
\begin{equation}
\begin{split}
p_i^{(n + 1/2)} &= p_i^{(n)} - \frac{\epsilon}{2} \frac{\partial U}{\partial q_i}(q^{(n)}) \\
q_i^{(n + 1)} &= q_i^{(n)} + \epsilon \frac{\partial K}{\partial p_i}(p^{(n + 1/2)}) \\
p_i^{(n + 1)} &= p_i^{(n + 1/2)} - \frac{\epsilon}{2} \frac{\partial U}{\partial q_i}(q^{(n + 1)})
\end{split}
\end{equation}
First a half-step for the momentum variables is computed, which is then used for a full position step. Finally, a second momentum half-step based on the updated position completes the leapfrog step. Since each of these updates is simply a shear transformation in $(q, p)$-space and therefore has a determinant of 1, a complete leapfrog step also has a determinant of 1 and is volume-conserving. If we perform multiple leapfrog steps, we can jump directly from $p_i^{(n + 1/2)}$ to $p_i^{(n + 3/2)}$ for greater efficiency.

With the usual choice for the kinetic energy $K(p) = p^T M^{-1} p /2$ and some manipulation of the above equations we can obtain an alternative formulation of the leapfrog method, which is more intuitive (but computationally more expensive):
\begin{equation}
\begin{split}
q^{(n + 1)} &= q^{(n)} + \epsilon M^{-1} p^{(n)}) + (\epsilon^2/2) M^{-1} F(q^{(n)}) \\
p^{(n + 1)} &= p^{(n)} + \epsilon (F(q^{(n)}) + F(q^{(n+1)}))/2,
\end{split}
\end{equation}
where $F(q) = - \nabla U(q)$ is the force acting on the particle at position $q$ due to the potential energy landscape. Since $M$ corresponds to the mass of the particle, $M^{-1} p$ gives its velocity and $M^{-1} F(q)$ its acceleration. From the first equation we see that the leapfrog method updates the position assuming motion under constant acceleration: $q(t) = q_0 + v_0 t + 1/2 a t^2$ with a initial position $q_0 = q^{(n)}$, initial velocity $v_0 = M^{-1} p^{(n)}$ and acceleration $a = M^{-1} F(q^{(n)})$. The second equation, which gives the momentum update, is simply a discretized version of the basic relationship $dp/dt = F$, i.e.\ force equals change of momentum, using the average of the forces at the start and the end point.

The local error of the leapfrog method, i.e.\ the error incurred in a single step, has order $\epsilon^3$; the global error, i.e.\ the error in the solution over a fixed time interval $L$, has order $\epsilon^2$. As a symplectic integrator the leapfrog method approximately conserves the Hamiltonian, so that the global error in the Hamiltonian, which is also order $\epsilon^2$, usually does not grow exponentially with the simulation length $L$ (with $\epsilon$ fixed) as it may for many other integration schemes \parencite{Neal2011}.

\subsection{The HMC algorithm}
\label{sec:HMCAlgorithmSection}
\subsubsection{Relating probability density to energy}
In order to apply HD within an MCMC method to sample from some target distribution, we need to derive appropriate energy functions. A key relationship in statistical mechanics is $f_S(s) \propto \exp \left(- E(s) \right)$, relating the probability density $f_S(s)$ for observing a particle in state $s$ with the energy $E(s)$ of that state.\footnote{Here, w.l.o.g., we set the temperature $T$ of the system to be the reciprocal of the Boltzmann constant.}. The distribution given by this probability density function is called the \textit{canonical distribution}.

By inverting this relationship we can derive the appropriate energy from any target distribution. The potential energy $U(q)$, whose canonical distribution has the target density $f_\textrm{target}(q)$, is thus given by $U(q) = -\log f_\textrm{target}(q)$, where we can drop any additive constant arising from the above proportionality relation, because energies only influence the particle motion through their derivatives. This also means that we do not need $f_\textrm{target}$ to be normalized. A closer look at $U(q)$ reveals that it equals the negative log-likelihood (NLL) of $f_\textrm{target}(q)$, which is frequently used as a minimization objective in machine learning. Therefore, this potential energy will promote motion towards low NLL points and thus the points proposed by motion simulation with this potential energy will tend to have a higher likelihood than those proposed by other methods.

For the simulation by HD the state of the system consists of the variable of interest $q$ plus an auxiliary momentum variable $p$ of the same size and so is given by the $2d$-dimensional $s = (q, p)$. With the potential energy $U(q)$ derived from the target distribution as described above, the Hamiltonian of this system is given by $H(q, p) = U(q) + K(p)$ for some kinetic energy $K(p)$ of our choice. Due to the additive nature of this Hamiltonian the joint canonical distribution of $(q, p)$ factorizes:
\begin{equation} \label{eq:JointDensity}
\begin{split}
p(q, p) &\propto \exp \left( -H(q, p) \right) \\
			&\propto f_\textrm{target}(q) \cdot \exp{(-K(p))}
\end{split}
\end{equation}

\subsubsection{Choice of kinetic energy}
In order to obtain a Markov chain, whose invariant distribution is the canonical distribution, some restrictions apply to the choice of kinetic energy \parencite{Betancourt2014}. In particular, the corresponding \textit{canonical momentum distribution} $f_\textrm{kin}(p) \propto \exp{(-K(p))}$ should have a mean of zero, since otherwise reversing the dynamics and computing the acceptance probability (detailed below) become unnecessarily complicated. While it is possible to make the kinetic energy dependent on position in the Riemann Manifold Hamiltonian Monte Carlo method \parencite{Girolami2011}, this requires complicated modifications to the integrator and will not be considered here. \textcite{Betancourt2014} argue that there is little motivation to choose a kinetic energy other than the quadratic form from classical physics and in the following we will assume the usual choice for the kinetic energy
\begin{equation} \label{eq:KineticEnergy}
K(p) = p^T M^{-1} p/2
\end{equation}
for some positive definite mass matrix $M$. The corresponding canonical momentum distribution (after normalization) $f_\textrm{kin}$ is the multivariate Gaussian distribution with mean zero and covariance matrix $M$.

\subsubsection{The algorithm}
The HMC algorithm (see algorithm~\ref{alg:HMC}) produces the desired Markov chain \parencite{Neal2011}. There are two main steps in the algorithm: Firstly the simulation of HD using a reversible and volume-preserving integrator, e.g. the leapfrog method, and secondly a Metropolis-Hastings acceptance step to ensure the desired invariant distribution. Due to the momentum negation of the proposed state in the third step of the algorithm, the proposal distribution is symmetrical because of the reversibility of the integration method. As a result $\tilde{q}(\tilde{s}_t|s_{t-1}) = \tilde{q}(s_{t-1}|\tilde{s}_t)$ holds in the Metropolis-Hastings acceptance probability in equation~\eqref{eq:Metropolis-Hastings}, so the acceptance probability simplifies to
\begin{equation} \label{eq:AcceptanceProbability}
p_{\textrm{accept}}(s^*_{t-1}) = \min[1, \exp(-H(\tilde{s}_t) + H(s^*_{t-1}))].
\end{equation}

\begin{algorithm}
\caption{The HMC algorithm}\label{alg:HMC}
\begin{algorithmic}[1]
\Require Numeric integrator $HD(s)$ of Hamilton's equations simulating HD starting from state $s$ for a fixed length
\Require Current state $s_{t-1} = (q_{t-1}, p_{t-1})$
\State Sample new momentum $p^*_{t-1}$ from $f_\textrm{kin}$
\State Simulate HD starting from $s^*_{t-1} = (q_{t-1}, p^*_{t-1})$
\State Negate the momentum of the resulting state $s_\textrm{HD} = HD(s^*_{t-1})$ to obtain the proposed state $\tilde{s}_t = (q_\textrm{HD}, - p_\textrm{HD})$
\State Compute the acceptance probability $p_\textrm{accept}=p_\textrm{accept}(s^*_{t-1})$ as defined by equation~\eqref{eq:AcceptanceProbability}
\State Accept the move from $s^*_{t-1}$ to $\tilde{s}_t$ with probability $p_\textrm{accept}$
\State \textbf{Return} new state $s_t$
\end{algorithmic}
\end{algorithm}

It can be shown that this algorithm conserves the canonical distribution, which therefore also is the invariant distribution of the constructed Markov chain \parencite{Neal2011}. If the HD simulation was exact, then the Hamiltonian would be conserved, since negation of the momentum does not change the value of the Hamiltonian due to its symmetry. Therefore the acceptance probability would always be 1. However, numeric integrators cannot conserve the Hamiltonian exactly, necessitating the acceptance step. Still, for symplectic integrators, such as the leapfrog method, the numerical error usually remains bounded, allowing the rejection rate to be kept small even for long simulations.

\begin{figure*}[t]
\centering
\includegraphics[width=2.05\columnwidth]{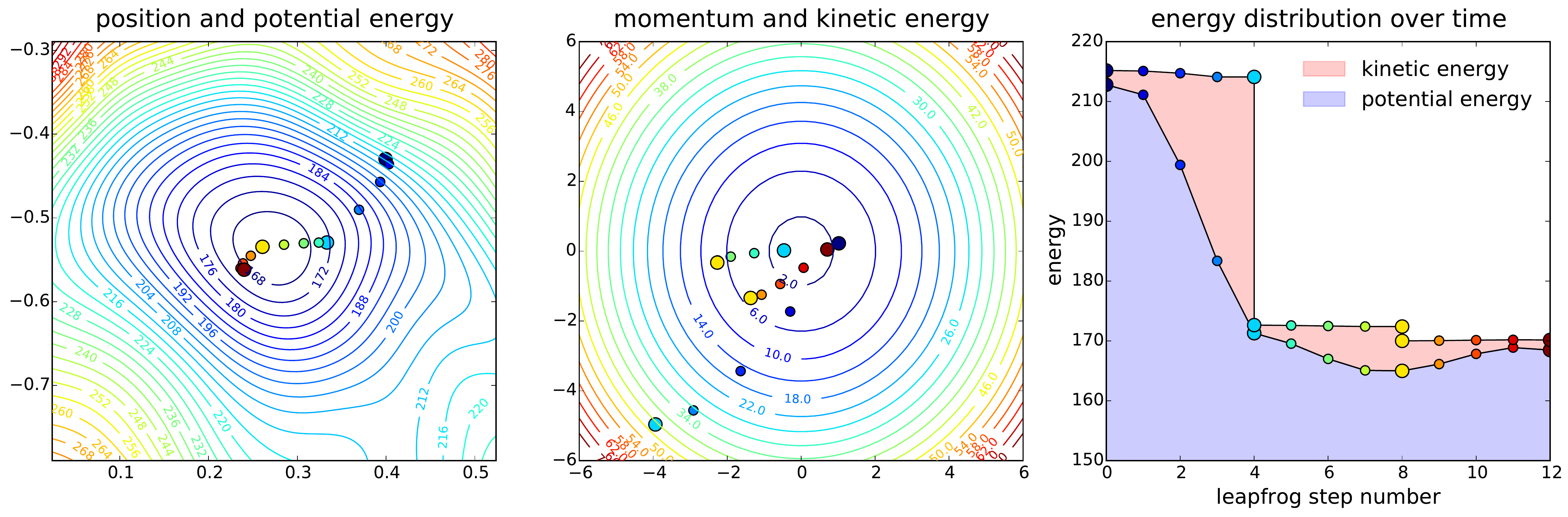}
\caption{Evolution of a particle under the HMC algorithm with 3 HMC steps consisting of 4 leapfrog steps each. Each computed point along the trajectory is indicated by a separate color ranging from dark blue (starting point) to dark red (final point). Thicker dots highlight points, where momentum resampling was performed. The left plot shows the position of the particle with the prescribed potential energy represented by the contour plot. The centre plot depicts the momentum of the particle with the kinetic energy at each point indicated by the contours. Where the momentum was resampled, two identical dots are shown for the state before and after the resampling. In the plot on the right the energy distribution of the particle over time is given, with the potential energy in blue and the kinetic energy in red.}
\label{fig:HMC_MOTION_3hmc_04lf}
\end{figure*}

Due to the (approximate) conservation of the Hamiltonian during HD, the joint density of $(q,p)$ given by~\eqref{eq:JointDensity} remains almost unchanged by steps 2 to 5 of the algorithm. Only the resampling of the momentum variable at the start of each HMC step allows large changes in the joint density. This can be seen in figure~\ref{fig:HMC_MOTION_3hmc_04lf}, where the evolution of a single particle is shown under the HMC algorithm. During the leapfrog steps the potential energy of the particle is partly converted to kinetic energy. With the newly drawn momentum the kinetic energy of the particle is smaller than before in this example leading to a decrease of its total energy. The sampled kinetic energy is given by $(1/2) p^T M^{-1} p$ (ignoring additive constants), which is $(1/2) \cdot \chi^2_d$-distributed for any $M$, if $p \sim f_\textrm{kin}(p)$. For the two-dimensional example in the figure this means that on average a particle gets a kinetic energy of 1 at the start of each HMC step, which could be converted into potential energy. Since the craters in the potential energy landscape are much deeper, particles are very unlikely to leave such a crater once they are caught inside.

\begin{figure*}[t]
\centering
\includegraphics[width=2.05\columnwidth]{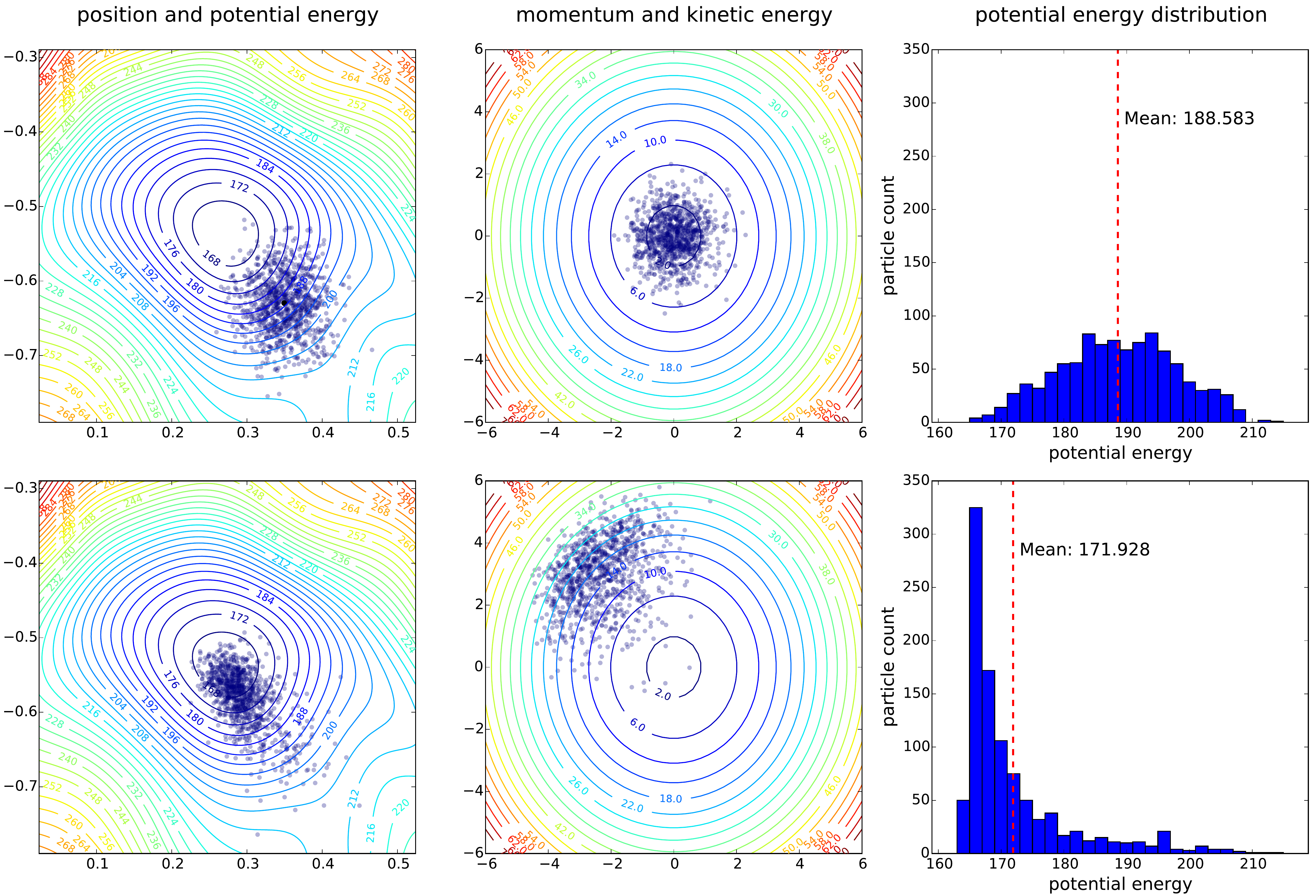}
\caption{Evolution of an ensemble of 1000 particles under the HMC algorithm: The first row of plots shows the initial state of the system and the second row the state after an HMC step. The plots on the left give the positions of the particles with the prescribed potential energy represented by the contour plot. The centre plots depict the arrival momenta of the particles with the kinetic energy indicated by the contours. The right-hand plots show histograms of the potential energies of the particles.}
\label{fig:HMC_Effect_Illustration}
\end{figure*}

Simulating an ensemble of particles illustrates how the convergence to the desired distribution happens in the HMC algorithm. In figure~\ref{fig:HMC_Effect_Illustration} particles were distributed according to some supposed distribution different from the desired distribution, which determines the energy landscape. After the first HMC step (bottom row of plots) the particles have mostly slid downhill, which can also be seen in the change in their potential energy (plots in the right column). Correspondingly, they have picked up kinetic energy, which will, however, be removed at the start of the next HMC step. In this way, the HMC steps initially reduce the amount of potential energy in the system corresponding to an increase of the likelihood of the particles w.r.t.\ the target distribution. By sampling a new momentum at the start of each HMC step, instead of for example setting it to 0 (in which case all the particles would gather at the low point of the potential energy), we ensure that the particles remain spread out and are eventually distributed according to the target distribution. 

\subsection{Effect of the kinetic energy covariance matrix}
\label{sec:EffectOfKineticEnergyChoice}

For simplicity we restrict the kinetic energy (see eq.~\eqref{eq:KineticEnergy}) to be a positive-definite quadratic form, but not necessarily with a scalar multiple of the identity as mass matrix as the physical intuition of particle mass would suggest. A possible interpretation of such a "mass" matrix would be that the inertial mass of the particle, i.e.\ its resistance to change in its velocity, is non-isotropic. In other words, the particle is more responsive to forces in some directions than in others. This somewhat non-physical freedom, however, has a very nice effect in the HMC algorithm: It allows an implicit rescaling of the $q$-space as explained below.

Such a rescaling can be very beneficial for the numerical solution, because the most restricted direction (with the most extreme changes in potential energy) limits the step length $\epsilon$ to be used in the discrete simulation. If a larger step length is used, the approximations of the energy surface used in the simulation are too coarse in the restricted direction and the discretization error becomes very large. As a result one may have to choose a very small step size, but this then limits the motion in the less restricted directions, where a larger step size would allow faster movement through the state space. Therefore, by rescaling the space we can achieve a more equal scaling in each direction, so that neither large errors nor slow exploration hamper the performance of the algorithm.

To see the connection between the mass matrix and the rescaling of $q$-space, assume the numerics of the dynamics w.r.t.\ the original variables $(q, p)$ were badly scaled when using the physically intuitive $K(p) = p^T p/2$ (taking $m=1$ for simplicity). Further suppose a transformation $q' = A^{-1} q$ with $p'=p$ and the same kinetic energy would yield a better scaling for some non-singular matrix $A$. Then the target distribution for $q'$ is given by $f_\textrm{target}'(q') = f_\textrm{target}(Aq')/|\det(A^{-1})|$ in terms of the original target distribution $f_\textrm{target}(q)$. Hence, the corresponding potential energy is $U'(q') = U(Aq')$, where we can drop the additive $\log(|\det(A^{-1})|)$ term. From Hamilton's equations~\eqref{eq:HamiltonsEquations} for this system we get the following equations for the motion in terms of the original variables $(q, p)$:
\begin{equation}
\begin{split}
\frac{dq}{dt} &= A \frac{dq'}{dt} = Ap' = Ap \\
\frac{dp}{dt} &= \frac{dp'}{dt} = - \nabla U'(q') = - A^T \nabla U(q)
\end{split}
\end{equation}
The evolution of the position variable $q$ is thus given by (compare Newton's equation of motion~\eqref{eq:NewtonsEquation}):
\begin{equation} \label{eq:EvolutionQTransformed}
\frac{d^2q}{dt^2} = A \frac{dp}{dt} = - A A^T \nabla U(q)
\end{equation}

Now alternatively, let us consider the untransformed system, but with the kinetic energy $K''(p) = p^T A A^T p$. Then Hamilton's equation give us:
\begin{equation}
\begin{split}
\frac{dq}{dt} &= A A^T p \\
\frac{dp}{dt} &= - \nabla U(q),
\end{split}
\end{equation}
which results in the same evolution of the variable of interest $q$ as the direct transformation of $q$ above (compare equation~\eqref{eq:EvolutionQTransformed}). Regarding the evolution of $q$ these two approaches are thus identical (although the $p$ trajectories differ).

Introducing this transformation via the kinetic energy rather than transforming $q$ directly has the advantage, that we do not manipulate the variables of interest, which may be needed in their original form. Instead, we can achieve the same rescaling by modifying the auxiliary momentum variables, which do not have any external significance.

\subsection{Partial momentum updates}
\label{sec:PartialMomentumUpdate}
If the number of leapfrog steps is small, subsequent points in the Markov chain generated by the HMC algorithm may be close to each other and highly correlated. This is especially obvious, if we imagine a flat plateau in the potential energy surface: Whatever momentum is sampled at the start of the HMC step, the simulated motion may frequently end at some other point still on the plateau, if the number of leapfrog steps is small. There the same may happen again, perhaps even bringing us back to the previous point, leading to an inefficient random-walk-like behaviour on this plateau.

To counter such a behaviour \textcite{Horowitz1991} proposed an extension to HMC, where the momentum is only partially updated. So instead of overwriting the momentum variable with a random sample from the canonical momentum distribution, the idea is to use a weighted sum of the current momentum and the newly drawn sample. By doing this the particle does not completely loose its current momentum after each HMC step, but continues in a similar direction as before. In the plateau example above, this means the particle is very unlikely to double back on its previous progress and will rather travel across the plateau in a directed fashion, avoiding the random-walk-like behaviour of the base HMC algorithm.

Some care must be taken in combining the current momentum $p_{t-1}$ with the new sample $p_\textrm{sampled}$, because this momentum scrambling step must conserve the canonical distribution. This can be done by defining the updated momentum $p^*_{t-1}$ by
\begin{equation} \label{eq:partialMomentumUpdate}
p^*_{t-1} = \alpha \cdot p_{t-1} + \sqrt{1 - \alpha ^2} \cdot p_\textrm{sampled}
\end{equation}
for some $\alpha \in [-1, 1]$. In the converged chain both $p_{t-1}$ and $p_\textrm{sampled}$ are distributed according to the canonical distribution (Gaussian with mean zero and covariance matrix $M$), so $p^*_{t-1}$ will also be Gaussian and have mean zero. Since $p_{t-1}$ and $p_\textrm{sampled}$ are also independent of each other, the covariance is $\Cov(p^*_{t-1}) = \alpha^2 \cdot M + (1 - \alpha ^2) \cdot M = M$ as required.

\begin{algorithm}
\caption{The HMC algorithm with partial momentum updates}\label{alg:HMCWithPartial}
\begin{algorithmic}[1]
\Require Numeric integrator $HD(s)$ of Hamilton's equations simulating HD starting from state $s$ for a fixed length
\Require Current state $s_{t-1} = (q_{t-1}, p_{t-1})$
\State Sample new momentum $p_\textrm{sampled}$ from $f_\textrm{kin}$
\State Update the momentum as in equation~\eqref{eq:partialMomentumUpdate} to obtain $p^*_{t-1}$
\State Simulate HD starting from $s^*_{t-1} = (q_{t-1}, p^*_{t-1})$
\State Negate the momentum of the resulting state $s_\textrm{HD} = HD(s^*_{t-1})$ to obtain the proposed state $\tilde{s}_t = (q_\textrm{HD}, - p_\textrm{HD})$
\State Compute the acceptance probability $p_\textrm{accept}=p_\textrm{accept}(s^*_{t-1})$ as defined by equation~\eqref{eq:AcceptanceProbability}
\State Accept the move from $s^*_{t-1}$ to $\tilde{s}_t$ with probability $p_\textrm{accept}$
\State Negate the momentum to obtain the new state $s_t$
\State \textbf{Return} new state $s_t$
\end{algorithmic}
\end{algorithm}

\begin{figure}
\centering
\includegraphics{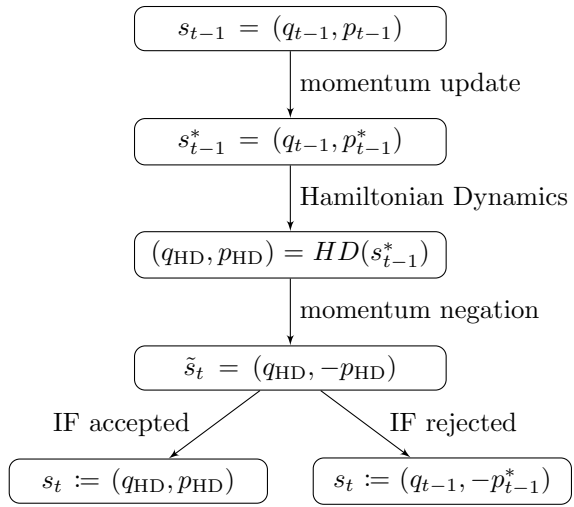}
\caption{Flow chart illustrating the steps of the HMC algorithm with partial momentum updates.}
\label{fig:HMC_schematic}
\end{figure}

Algorithm~\ref{alg:HMCWithPartial} shows the steps in the improved version of the HMC algorithm for generating the next state of the Markov chain. Like the original HMC algorithm (algorithm~\ref{alg:HMC}), which can be recovered by setting $\alpha = 0$, this extension preserves the joint canonical distribution and thus yields a Markov chain with the required properties. Step 7, which is missing in the base version, is important for the case with partial momentum updates: If the proposed state was accepted, this step reverses the earlier momentum negation so that the particle keeps its direction. If the proposal was rejected, then it flips the momentum and the particle doubles back on itself. This can be clarified by combining steps 6 and 7:
\begin{equation} \label{eq:StateAfterAcceptReject}
s_t := \begin{cases} s_\textrm{HD} & \textrm{if accepted} \\ 
								(q_{t-1}, -p^*_{t-1}) & \textrm{if rejected}
					  \end{cases}
\end{equation}
For a better understanding, the order and used nomenclature of the states, which will be needed to derive the variational lower bound in section~\ref{sec:HMCVI}, are illustrated in figure~\ref{fig:HMC_schematic}.

While the partial momentum update brings little benefit, if the number of leapfrog steps is large, it was reported to be beneficial for chains with shorter-than-optimal trajectories \parencite{Neal2011}. Because of computational limitations this will usually be the case in our application of HMC.
\section{Variational inference with HMC}
\label{sec:HMCVI}
As suggested by \textcite{Salimans2014} HMC is a very good MCMC method to be used within MCVI as introduced in section~\ref{sec:MCVI}, because it is very efficient, usually requiring fewer steps than other methods for good convergence. However, some care must be taken in the derivation of the auxiliary lower bound, since now the state of the generated Markov chain is not just the variable of interest $z$, but also the auxiliary momentum variable, which we will call $v$ (as it is related to velocity). The complete state is thus given by the $2d$-dimensional $s=(z, v)$, corresponding to the state $(q, p)$ in the previous section. The appropriate potential energy $U(z)$ is derived from the posterior density $p(z|x) \propto p(x, z)$, which is known upto a multiplicative constant from Bayes' Theorem:
\begin{equation} \label{eq:VIwithHMCPotEnergy}
U(z) = -\log p(x, z).
\end{equation}

Unless stated otherwise, the results below will hold for the more general algorithm with partial momentum updates, from which the standard HMC algorithm can be recovered by setting $\alpha = 0$. For notational ease we will write $u_{t-1}$ for the updated momentum, which was referred to as $p^*_{t-1}$ in the previous section. 

For the initial state of the chain we sample the position from a parametric approximation $q_0(z_0|x)$ and the momentum from the distribution $f_\textrm{kin}(v_0|x)$ corresponding to the chosen kinetic energy, so the density of the initial state is $q_0(s_0|x)=q_0(z_0|x) \cdot f_\textrm{kin}(v_0|x)$. Interestingly, there is no theoretical reason for the kinetic energy to be independent of $x$. This can be exploited to improve the quality of the bound (see section~\ref{sec:MassMatrixChoice} below). \label{sec:KinEnergyMayDependOnX}

\subsection{Deriving the variational lower bound}

The auxiliary lower bound given in equation~\eqref{eq:MCVIAuxLowerBound} can not be used with the HMC algorithm, since there the transition density $q(z_t|z_{t-1}, x)$ is intractable. The transition densities $q(s_t|s_{t-1}, x)$, however, can be easily computed (shown below). To incorporate these, the derivation of the auxiliary lower bound must be modified:\begin{equation}
\begin{split}
\log & p(x) \geq \mathcal{L} \\
& \geq \mathcal{L} - \E_{q(z_T|x)} \big[ D_{KL}[q(y | z_T, x) || r(y | z_T, x)] \big] \\
& = \E_{q(y, z_T|x)} \Big[ \log p(x, z_T) - \log q(y, z_T|x) \\
& 	\qquad\qquad\qquad\qquad 							+ \log r(y | z_T, x) \Big] \eqqcolon \mathcal{L}_{\textrm{aux}},
\end{split}
\end{equation}
where $y = (s_0, \dots, s_{T-1}, v_T)$.

Using the Markov property the density of the forward chain can be decomposed into the tractable transition densities and the density of the initial state: $\log q(y, z_T|x) = \log q_0(s_0|x) + \sum_{t=1}^T \log q(s_t|s_{t-1}, x)$. For the auxiliary reverse density we can rewrite $r(y | z_T, x) = r(s_0, \ldots, s_{T-1}| s_T, x) \cdot r_{\textrm{final}}(v_T | z_T, x)$ for some distribution $r_{\textrm{final}}(v_T|z_T, x)$, which approximates the final distribution of the momentum $v_T$ given the position $z_T$. By then assuming a Markov structure on the reverse model (as for the base case) we get $\log r(y | z_T, x) = \log r_{\textrm{final}}(v_T | z_T, x) + \sum_{t=1}^T \log r(s_{t-1}|s_t, t, x)$, where the reverse model $r$ may depend on the time step (as discussed in section~\ref{sec:MCVI}). With these assumptions we can rewrite the lower bound as
\begin{equation} \label{eq:HMCVIAuxLowerBound}
\begin{split}
&\mathcal{L_\textrm{aux}} = \E_{q(s_0, \ldots, s_T|x)} \Big[ \log p(x, z_T) - \log q_0(z_0|x) \\
&\;+ \log r_{\textrm{final}}(v_T | z_T, x) - \log f_\textrm{kin}(v_0|x)  \\ 
&\;+ \sum \limits_{t=1}^T \big( \log r(s_{t-1}|s_t, t, x) - \log q(s_t|s_{t-1}, x) \big) \Big].
\end{split}
\end{equation}
For this bound auxiliary models must be learnt for the reverse transition model $r(s_{t-1}|s_t, t, x)$ and for $r_{\textrm{final}}(v_T | z_T, x)$, which we will refer to as the final momentum model. Additionally, we can learn the step size $\epsilon$ and the covariance matrix (or mass matrix) $M$ of the kinetic energy used by the HMC algorithm. The number of HMC steps and the number of leapfrog steps per iteration have to be integer and are therefore complicated to learn. For this reason, they will be considered as hyperparameters of the algorithm, which are fixed in advance. Optimization of this bound is done as for MCVI (compare section~\ref{sec:MCVI}) by using Monte Carlo estimates of the expectation of the gradient. For future reference we will call the optimization of this lower bound Hamiltonian Monte Carlo Variational Inference (HMCVI).

To evaluate this lower bound, the transition probabilities $q(s_t|s_{t-1}, x)$ implied by the HMC algorithm must be computed. A key observation here is that performing HD on the variables with a volume-preserving integrator, such as the leapfrog method, is a bijective and volume-preserving mapping. Therefore, the change of variables\footnote[1]{The density after the change of variables will be marked by an apostrophe, since it formally is a different function.} between the proposed state $\tilde{s}_t = (\tilde{z}_t, \tilde{v}_t)$ and the state $s^*_{t-1} = (z_{t-1}, u_{t-1})$ from which the HD simulation was started, is bijective and has a Jacobian determinant equal to 1 (see figure~\ref{fig:HMC_schematic} for the used naming of intermediate states in the HMC algorithm). In the following, we will write $revHD(s)$ to denote the state which results from running HD backwards in time starting from $s$. Further $\delta[.]$ will be used to signify the Dirac $\delta$-function.

\subsection{Transition densities without the acceptance step}
\label{sec:TransitionDensitiesNoAcceptance}
If we leave out the acceptance step in the HMC algorithm, the proposed state is always accepted as the new state, so $s_t = \tilde{s}_t$. In this case, the transition densities of the forward model follow directly from the bijectivity and volume-preservation of HD:
\begin{equation} \label{eq:ForwardTransitionNoAcceptance}
\begin{split}
q(s_t|&s_{t-1}, x) = q'(revHD(s_t)|s_{t-1}, x) \\
&= q'(z^*_{t-1}, u_{t-1} |z_{t-1}, v_{t-1}, x) \\
&= q_U(u_{t-1}|v_{t-1}, x) \cdot \delta[{z}^*_{t-1} - z_{t-1}],
\end{split}
\end{equation}
where $(z^*_{t-1}, u_{t-1}) \coloneqq revHD(z_t, v_t)$. With $v_{\textrm{samp}} \coloneqq (u_{t-1} - \alpha \cdot v_{t-1})/{\sqrt{1 - \alpha^2}}$, the momentum drawn from the canonical momentum distribution in this step, we can simplify the density of the updated momentum
\begin{equation} \label{eq:qUDefinition}
q_U(u_{t-1}|v_{t-1}, x) = f_\textrm{kin}(v_{\textrm{samp}}|x) \cdot (\frac{1}{\sqrt{1 - \alpha^2}})^{d}.
\end{equation}

For the reverse model $r(s_{t-1}|s_t, t, x)$ we can also exploit the properties of HD to simplify the model to be learnt (with the same notation):
\begin{equation} \label{eq:ReverseTransitionNoAcceptance}
\begin{split}
r(&s_{t-1}|s_{t}, t, x) = r'(z_{t-1}, v_{t-1} |z^*_{t-1}, u_{t-1},t , x) \\
&\;= r_V(v_{t-1}|z_{t-1}, u_{t-1}, t, x) \cdot \delta[z^*_{t-1} - z_{t-1}] \\
\end{split}
\end{equation}
Thus the auxiliary reverse model is fixed except for the density $r_V$ of the \textit{arrival} momentum $v_{t-1}$, with which the position $z_{t-1}$ was reached. As inputs to a model of this distribution we may use the position $z_{t-1}$, $x$, the current time step $t$ and the updated momentum $u_{t-1}$, with which the particle left the position $z_{t-1}$. All of these may contain information about the arrival momentum, so they all should be included for a better fitting model.

For the computation of the lower bound the Dirac $\delta$-functions are problematic, because their value is infinite, when their argument equals $0$. However, since a $\delta$-function appears both in the forward and in the reverse model (whose log-likelihoods are subtracted from each other), the $\delta$-functions can be handled: $\delta(x)$ can be approximated by a function with an extended support of width $\kappa$, where its value is $1/\kappa \cdot \mathbb{I}\big[x \in (x - 0.5 \cdot \kappa, x + 0.5 \cdot \kappa)\big]$. Here, $\mathbb{I}[x \in A]$ denotes the indicator function of some set $A$, which equals $1$, if $x \in A$, and is $0$ otherwise. Like for the $\delta$-function, the integral of this approximation over the real line is 1. Therefore taking the limit of this approximation as $\kappa \rightarrow 0$ gives $\delta(x)$. When subtracting the logarithms of two such approximations, the $1/\kappa$ factors cancel, so we can safely take the limit and are left with two indicator functions instead of the two Dirac $\delta$-functions.

The main drawback of leaving out the acceptance step is that the canonical distribution of the state is no longer preserved by the Markov chain transitions and as a result the chain does no longer converge to the canonical distribution. This means that samples from the converged chain will not follow the target distribution. While this would rule out the algorithm for its usual sampling application, it may still be of use for improving the approximation of the posterior distribution, because here it is usually only feasible to perform a very limited number of HMC steps for computational reasons. Thus loosing the asymptotic convergence is acceptable, since the initial steps of the chain should be similar. Apart from the computational simplifications, leaving out the acceptance step also makes the algorithm less wasteful, since no proposals are discarded.

\subsection{Transition densities with the acceptance step}

When using the latent variable $z$ alone as the state, the transition density $q(z_{t-1}|z_{t-1},x)$ for staying at the same location cannot be computed for the Metropolis-Hastings algorithm: Either the proposed state exactly matched the old state or a now unknown proposed state was rejected. Computing the probability of the second possibility requires the integration of the rejection probability over all possible proposed points, i.e.\ the integral $\int_{z} \tilde{q}(z|z_{t-1}, x) \cdot (1 - p_{\textrm{accept}}(z_{t-1}, z)) dz$, where $\tilde{q}$ is the proposal density and $p_{\textrm{accept}}$ is the acceptance probability defined in equation~\eqref{eq:Metropolis-Hastings}. This is usually intractable. A possible solution would be to explicitly include a binary random variable in the state, which records the acceptance of the previous step. However, this would lead to non-differentiability of the lower bound \parencite{Salimans2014}.

Exploiting the structure of HMC, we can bypass this problem and include the acceptance step without introducing any new variables, because in case of rejection the momentum variable is not reset to its previous value, but keeps the updated value (see equation~\eqref{eq:StateAfterAcceptReject} in section~\ref{sec:PartialMomentumUpdate}). In this way it stores the proposed state, which was rejected. This removes the problematic integral and thus makes the transition density tractable, as we will demonstrate in detail below.

Crucially, by including the acceptance step in the algorithm, convergence of the Markov chain to the true posterior is guaranteed. Hence, an arbitrarily exact approximation to the posterior can be obtained by performing a sufficient number of HMC steps.

\subsubsection{Forward model}

For the derivation of the transition density $q(s_t|s_{t-1}, x)$, let $A$ be the random variable indicating, whether the proposed move was accepted or not, i.e.\ $A=1$, if the move was accepted, and $A=0$ otherwise. For extra clarity, we will in the following write out the probability density functions (denoted by $f$) with the variables explicitly given in the subscript, so for example $q(s_t|s_{t-1}, x) = f_{S_t|S_{t-1}, X}(s_t|s_{t-1}, x)$. Using the law of total probability we can then decompose the transition density as follows:
\begin{equation}
\begin{split}
&f_{S_t|S_{t-1}, X}(s_t|s_{t-1}, x)\\
&\;=\sum_{a=0}^1 \int f_{S_t, A, U_{t-1}|S_{t-1}, X}(s_t, a, u|s_{t-1}, x) du \\
&\;=\sum_{a=0}^1 \int f_{S_t|A, U_{t-1}, S_{t-1}, X}(s_t| a, u, s_{t-1}, x) \\
&\qquad\qquad\cdot \mathbb{P}(A = a|U_{t-1} = u, S_{t-1} = s_{t-1}, x) \\
&\qquad\qquad \cdot f_{U_{t-1}|S_{t-1}, X}(u|s_{t-1}, x) du
\end{split}
\end{equation}
Each term in this expression can be computed:
\begin{itemize}
\item $f_{U_{t-1}|S_{t-1}, X}(u|s_{t-1}, x) = q_U(u|v_{t-1}, x)$ as in equation~\eqref{eq:qUDefinition} for the forward transition without the acceptance step.
\item $\mathbb{P}(A=1|S_{t-1} = s_{t-1}, U_{t-1} = u, x) = p_{\textrm{accept}}(z_{t-1}, u)$ as in equation~\eqref{eq:AcceptanceProbability} and correspondingly $\mathbb{P}(A=0|S_{t-1} = s_{t-1}, U_{t-1} = u, x) = 1- p_{\textrm{accept}}(z_{t-1}, u)$.
\item If the updated state $S^*_{t-1} = (Z_{t-1}, U_{t-1})$ and $A$ are known, the new state is uniquely determined, so $f_{S_t|A, U_{t-1}, S_{t-1}, X} = f_{S_t|A, S^*_{t-1}}$ with $f_{S_t|A, S^*_{t-1}}(s_t| 1, (z_{t-1}, u)) = \delta \left[s_t - HD(z_{t-1}, u) \right]$ and $f_{S_t|A, S^*_{t-1}}(s_t| 0, (z_{t-1}, u)) = \delta \left[s_t - (z_{t-1}, -u) \right]$.
\end{itemize}

Inserting these terms in the above decomposition and integrating out the delta functions gives
\begin{equation}
\begin{split}
q(s_t& |s_{t-1}, x) \\
&= \delta \left[z_{\textrm{revHD}} - z_{t-1} \right] \cdot p_{\textrm{accept}}(z_{t-1}, v_{revHD}) \\
&\qquad\qquad\qquad \cdot q_U(v_{revHD}|v_{t-1}, x) \\
&\quad + \delta \left[ z_t - z_{t-1} \right] \cdot (1 - p_{\textrm{accept}}(z_{t-1}, -v_t)) \\
&\qquad\qquad\qquad \cdot q_U(-v_t|v_{t-1}, x),
\end{split}
\end{equation}
where we write $z_{\textrm{revHD}}$ and $v_{\textrm{revHD}}$ for the projections of $revHD(z_t, v_t)$ into $z$- and $v$-space respectively.

As we will see below, the reverse model densities will also contain a $d$-dimensional Dirac $\delta$-function in each summand, so we can apply the trick introduced in section~\ref{sec:TransitionDensitiesNoAcceptance} to replace $\delta$-functions by indicator functions. Here, the indicator functions can be taken to indicate, whether the proposed state was accepted (in the first summand) or rejected (in the second), because the probability of exactly achieving the equality inside the $\delta$-function in the opposite case is negligible, i.e.\ if the move is accepted, $z_t = z_{t-1}$ will not occur in practice. In the following, we will write $\mathbb{I}_\textrm{acc}$ for this indicator.

Thus, we can regard each summand as treating one of the acceptance/rejection cases. Writing $u_{t-1}$ for the updated momentum generated in the HMC algorithm, we have $u_{t-1} = v_{revHD}$ in the first summand and $u_{t-1} = -v_t$ in the second summand. In other words, the $q_U$ term in both summands is $q_U(u_{t-1}|v_{t-1}, x)$. Also, the $p_\textrm{accept}$ term is always computed from $u_{t-1}$ and the current position $z_{t-1}$, so the transition density can easily be calculated during the sampling process as
\begin{align}
\begin{split}
&q(s_t|s_{t-1}, x) = q_U(u_{t-1}|v_{t-1}, x) \\
&\;\;\;\cdot \Big( \mathbb{I}_\textrm{acc} \cdot p_{\textrm{accept}} + (1 - \mathbb{I}_\textrm{acc}) \cdot (1- p_{\textrm{accept}}) \Big),
\end{split}
\end{align}
where we can also simplify $q_U(u_{t-1}|v_{t-1}, x)$ as in equation~\eqref{eq:qUDefinition}.

\subsubsection{Reverse model}
\label{sec:TransDensitiesWithAcceptReverse}
In the lower bound we also need a density approximation for moves backwards through the chain, i.e.\ for $r(s_{t-1}|s_t, t, x) = f_{S_{t-1}|S_t,T, X}(s_{t-1} | s_t, t, x)$. By again letting $A$ be the event of accepting the proposed transition, we can apply the law of total probability to simplify the problem:
\begin{equation}
\begin{split}
&f_{S_{t-1}|S_t, T, X}(s_{t-1} | s_t, t, x) \\
&\qquad = \sum_{a} f_{S_{t-1}, A|S_t, T, X}(s_{t-1}, a | s_t, t, x) \\
&\qquad = \sum_{a} f_{S_{t-1} |A, S_t, T, X}(s_{t-1} | a, s_t, t, x) \\
&\qquad\qquad\qquad\qquad \cdot \mathbb{P}(A = a | S_t = s_t, t, x)
\end{split}
\end{equation}
The individual terms are now easier to handle:
\begin{itemize}
\item If we know that the previous move was accepted, we can use the reversibility of HD to obtain the state $S_{t-1}^* = (Z_{t-1}, U_{t-1})$, from which the HD-simulation was started, so
\begin{equation}
\begin{split}
&f_{S_{t-1} |A, S_t, T, X}(s_{t-1} | 1, s_t, t, x) \\
&\quad= f_{S_{t-1} |S_{t-1}^*, T, X}\big(s_{t-1} | revHD(s_t), t, x\big) \\
&\quad= \delta \left[z_{t-1} - z_{\textrm{revHD}} \right] \\
&\quad\qquad \cdot r_V(v_{t-1}|z_{\textrm{revHD}}, v_{\textrm{revHD}}, t, x) \\
&\quad= \mathbb{I}_\textrm{acc} \cdot r_V(v_{t-1}|z_{t-1}, u_{t-1}, t, x) 
\end{split}
\end{equation}
with $r_V$ as in equation~\eqref{eq:ReverseTransitionNoAcceptance} and the updated momentum $u_{t-1} = v_{\textrm{revHD}}$. As described earlier, the $\delta$-function is replaced by an indicator function by cancelling it against the $\delta$-functions in the forward density.
\item If the previous move was rejected, we know that the current state equals the state $S_{t-1}^*$ (with the momentum negated), so
\begin{equation}
\begin{split}
&f_{S_{t-1} |A, S_t, T, X}(s_{t-1} | 0, s_t, t, x) \\
&\quad= f_{S_{t-1} |S_{t-1}^*, T, X}\big(s_{t-1} | (z_t, -v_t), t, x\big) \\
&\quad= \delta \left[z_{t-1} - z_{t} \right] \cdot r_V(v_{t-1}|z_{t-1}, -v_t, t, x) \\
&\quad= (1 - \mathbb{I}_\textrm{acc}) \cdot r_V(v_{t-1}| z_{t-1}, u_{t-1}, t, x)
\end{split}
\end{equation}
where now $u_{t-1} = -v_t$ and the $\delta$-function is again converted to an indicator function by cancellation.

If these densities are computed during the sampling process, $u_{t-1}$ is directly available and does not need to be recomputed.
\item The probability $\mathbb{P}(A = 1|S_t = s_t, t, x)$ of accepting the previous step can be simplified under certain conditions (for the derivation see appendix~\ref{app:DerivationOfReverseAcceptanceProbability}): If $H(revHD(s_t)) \leq H(s_t)$, then $\mathbb{P}(A = 1|S_t = s_t, t, x) = 1$. Otherwise, this reverse acceptance probability needs to be learnt, but will tend towards $\exp(-H(revHD(s_t)) + H(s_t))$ as the chain converges. 
\item $\mathbb{P}(A = 0|S_t = s_t, t, x) = 1 - \mathbb{P}(A = 1|S_t = s_t, t, x)$
\end{itemize}

To capture the density of the backward Markov chain, a full auxiliary reverse model should therefore consist of two parts: Firstly the density estimating model for $r_V(v_{t-1}|z_{t-1}, u_{t-1}, t, x)$ as for the case without the acceptance step and secondly a model for $\mathbb{P}(A = 1|S_t = s_t, t, x)$. Regarding $r_V$, a small difference to the case without the acceptance step is that here $v_{t-1}$ is not always the end of a previous HD simulation, but can also be equal to  $-u_{t-2}$, the updated momentum at the start of the previous simulation, if the resulting proposal was rejected. 

Putting these terms together the reverse transition density is given by
\begin{equation}
\begin{split}
r(s_{t-1} &| s_t, t, x) = r_V(v_{t-1}| z_{t-1}, u_{t-1}, t, x) \\
&\cdot \Big( \mathbb{I}_\textrm{acc} \cdot \mathbb{P}(A = 1 | s_{t}, t, x) \\
&\qquad\; + (1 - \mathbb{I}_\textrm{acc}) \cdot \mathbb{P}(A = 0 | s_{t}, t, x) \Big)
\end{split}
\end{equation}
With this last component for the computation of the auxiliary lower bound, we are now able to apply the full HMC algorithm within the MCVI framework. In particular, we recover the guaranteed convergence to the exact posterior, which was lost by skipping the acceptance step.

\subsection{Learning the mass matrix}
\label{sec:MassMatrixChoice}
In its usual application as a sampling algorithm, the freedom in the configuration of the HMC is often a curse, since a lot of parameters have to be specified, for example the mass matrix and the step size. These choices may then dramatically change the performance of the algorithm. In our application, however, we can side-step this issue by allowing all continuous parameters of the algorithm to be learnt, in particular the mass matrix $M$. As explained in section~\ref{sec:EffectOfKineticEnergyChoice}, choosing a specific mass matrix is equivalent to a rescaling of the $z$-space, which may improve the convergence of the algorithm. It is important to keep in mind, that the space is not actually transformed, but that the mass matrix makes the algorithm behave as if the space was transformed.

In addition to this indirect contribution to the lower bound through improved convergence, the mass matrix also directly appears in the lower bound as the covariance matrix of the canonical momentum distribution. From the lower bound and the transition densities derived in the previous sections we see that for each HMC step a term $-\log f_\textrm{kin}(v_\textrm{samp}|x)$ appears in the bound. $f_\textrm{kin}$ is the density of the canonical momentum distribution, a zero-mean multivariate normal distribution with covariance matrix $M$, and $v_\textrm{samp}$ is a sample from this distribution. In the lower bound the expectation of this term is taken, so the contribution to the lower bound is
\begin{equation}
\begin{split}
&\E_{f_\textrm{kin}(v|x)} \Big[ -\log f_\textrm{kin}(v|x) \Big] \\
&\quad= \frac{1}{2}\E_{f_\textrm{kin}} \left[d\log(2 \pi) + \log(|M|) +  v^T M^{-1} v \right] \\
&\quad= \frac{1}{2} \Big( d \log(2 \pi) + \log(|M|) + d \Big), 
\end{split}
\end{equation}
since $v^T M^{-1} v$ has a $\chi^2$-distribution on $d$ degrees of freedom, which therefore has expected value $d$. 

%
In the reverse model we have the density $r_V(v_{t-1}|z_{t-1}, u_{t-1}, t, x)$ capturing the distribution of the arrival momentum. In other words, this tries to learn the momentum distribution at the end of the HD simulations. Thus, it should be closely related to the momentum distribution at the start of the HD simulations, which is exactly $f_\textrm{kin}$. In particular when assuming a multivariate normal density for $r_V$, their covariance matrices should be similar, so their direct contributions to the lower bound via forward and reverse densities should offset each other and not have a significant influence on the training of $M$.

The straight forward approach for the choice of mass matrix is to learn a single global mass matrix, which is used for all observed variables $x$. This corresponds to a global rescaling of the latent space for all computations within the algorithm. However, the potential energy $U(z)$ defining the landscape on which the dynamics are simulated, may strongly depend on $x$ (see equation~\eqref{eq:VIwithHMCPotEnergy}) and require a different rescaling for each $x$ for optimal performance. Therefore, a global rescaling will probably only have limited effect on the lower bound.

The obvious consequence of these considerations is to make the mass matrix dependent on $x$, which from a physical point of view corresponds to the masses of the simulated particles depending on the observed variable. This extension, which does not violate any theoretical considerations (see section~\ref{sec:KinEnergyMayDependOnX}), allows the optimal rescaling for each data point to be learnt and should greatly enhance the performance of the algorithm.

\subsection{Computational simplifications}

So far we have presented the theory behind HMCVI with the goal of mathematical completeness and clarity, but for an efficient implementation some simplifications can be made.

\subsubsection{Simplifications for HMCVI without partial momentum updates}
\label{sec:SimplificationWithoutPartialMomentumUpdate}
If we do not perform partial momentum updates, then the initial momentum $v_0$ is immediately replaced in the first step of the HMC algorithm. Thus, it should not influence the lower bound at all. And indeed, if $\alpha = 0$, $r(v_0|z_0, u_0, t=1, x) = f_\textrm{kin}(v_0|x)$ is the optimal choice for $r$ if $t=1$, since no more information about $v_0$ is available. In the loss only these terms contain $v_0$ and they appear with opposite sign in the loss, so by simply cancelling them instead of learning their equality we can reduce the computational load.

Furthermore, without partial momentum updates the updated momentum $u_t$ is directly sampled from the canonical momentum distribution, so it does not contain any information about the previous momentum $v_t$. Therefore, $u_t$ should not be used as an input in any of the reverse models, if $\alpha=0$. Conveniently, in this case the density predicting the arrival momentum $r_V$ has the same inputs as the final momentum model $r_\textrm{final}$, so we can combine them by setting $r_\textrm{final}(v_T | z_T, x) = r_V(v_T | z_T, T+1, x)$.

\subsubsection{Computing expectations explicitly}

The lower bound $\mathcal{L_\textrm{aux}}$ is given as the expectation of a sum of terms in equation~\eqref{eq:HMCVIAuxLowerBound}, but for some of these terms the expectation can be computed explicitly, reducing the noise in the stochastic gradient estimates used for training. In particular, the forward model density terms can usually be solved analytically, because the expectation over the sampled paths is actually the expectation over all the random variables determining this path. These random variables are the initial state sampled from $q_0$ and the various momentum updates all sampled from the canonical momentum distribution $f_\textrm{kin}$. For each of these variables the NLL appears as part of the lower bound and the expectation of the NLL of a random variable is actually its entropy, which is known in closed form for most distributions.
\section{Experimental results}
\label{sec:Experiments}
\subsection{Variational auto-encoders}

A very interesting and powerful application of VI is the so-called Variational Auto-Encoder (VAE), which was introduced by \textcite{Kingma2014} and \textcite{Rezende2014} independently. VAEs are used to estimate the probability density of a set of observations $\{x_i\}_{i=1, \dots, N}$ by assuming the existence of a more concise latent representation or \textit{encoding} $z_i$ for each observed point. This model can be trained by optimizing the lower bound $\mathcal{L}$ on the marginal likelihood $p(x_i)$ not only w.r.t.\ the parameters of the posterior approximation, but also w.r.t.\ the parameters of a generative model for $p(x_i, z_i)$ at the same time. Here, the generative model usually consists of a fixed prior for the latent variables $\pi(z_i)$ and a conditional distribution or \textit{decoder} $p(x_i|z_i)$ to be learnt. Correspondingly, the posterior approximation $q(z_i|x_i)$ is referred to as the \textit{encoder}.

In the following, we apply HMCVI to this model by enhancing the encoder through the addition of HMC steps and maximizing the auxiliary lower bound $\mathcal{L_\textrm{aux}}$. This should lead to an encoding closer to the best possible encoding given by the true but intractable posterior $p(z_i|x_i)$. In the HMC steps the generative model induces the energy surface on which the motion of particles is simulated. Therefore, in order to avoid numerical instabilities and unexpected behaviour, it is recommended to choose $p(x|z)$ to be smooth .

\subsection{The dataset and the effects of data binarization}
\label{sec:Dataset}
A common benchmark dataset for machine learning problems is the MNIST dataset compiled by \textcite{LeCun1998}, which consists of a total of 70000 $28 \times 28$ pixel images of handwritten digits. The usual modelling approach for probability density estimation of these images is to assume that the pixels follow Bernoulli distributions, so that sampled images are binary, i.e.\ only contain the values 0 (black) and 1 (white). However, while the underlying images were binary, the images in the dataset contain grey-scales due to the anti-aliasing techniques applied during the normalization preprocessing. To deal with this gap between the binary bi-level modelling approach and the smoother multilevel dataset, several strategies are in use.

\begin{figure}
\centering
\includegraphics[width=\columnwidth]{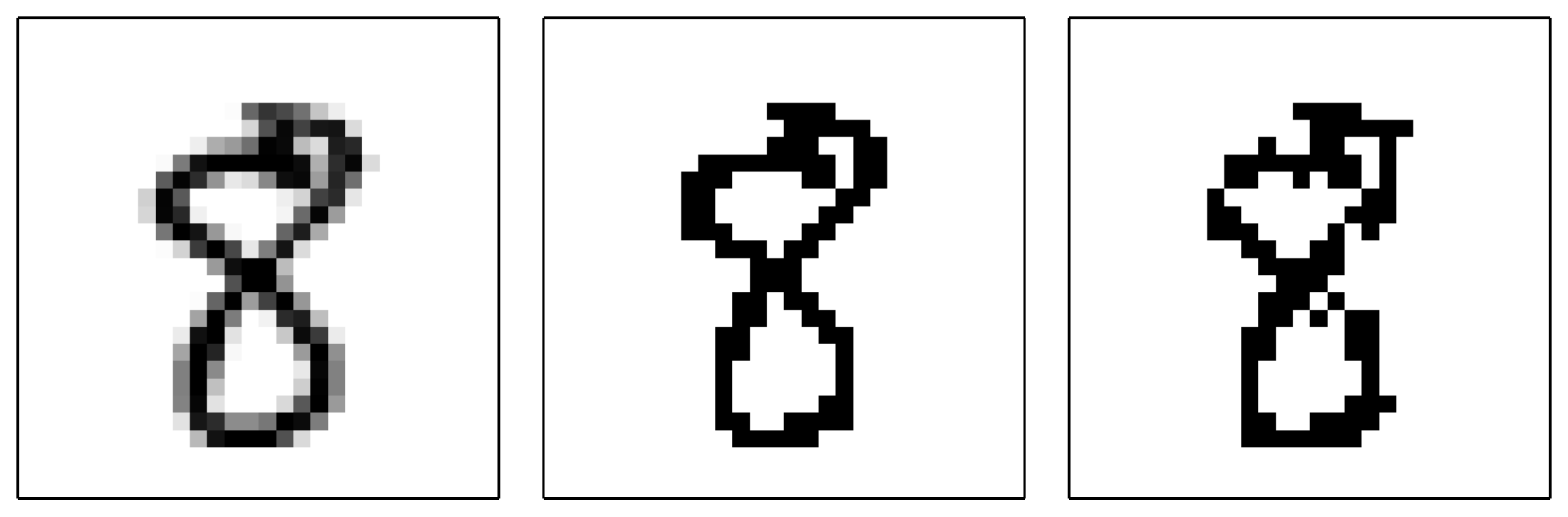}
\caption{Comparison of different binarization strategies on MNIST. The original (left) containing grey-scales was binarized using thresholding (middle) and stochastic binarization (right).}
\label{fig:MNISTBinarizationComparison}
\end{figure}

The most obvious approach is to directly use the unbinarized original dataset (fig.~\ref{fig:MNISTBinarizationComparison}, left), where pixel values range from $0$ to $255/256$ (with $256$ levels). A drawback of this method is its incompatibility with the assumption of a Bernoulli distribution, which leads to a lower likelihood of the model.
To avoid this incompatibility, it is necessary to binarize the images in the dataset. One way to do this is by applying a threshold to the pixel values, so setting the pixel to $1$, if its value is greater or equal to $0.5$, and to $0$ otherwise. This results in very clear images (fig.~\ref{fig:MNISTBinarizationComparison}, middle) and correspondingly a extremely high likelihood for most models. Although this is a very intuitive binarization strategy, it is rarely used in practice.

The most common binarization strategy for MNIST is stochastic binarization, which was introduced by \textcite{Salakhutdinov2008} and has become a standard benchmark for density estimation algorithms \parencite{Salimans2014,Rezende2014,Gregor2015}. Here, each pixel is randomly set to $1$ with the probability given by its value and to $0$ otherwise, so that taking the average over many draws from the same image returns the original unbinarized image. This procedure can produce somewhat unrealistic digits, for example with gaps, but still the digits are clearly recognizable (fig.~\ref{fig:MNISTBinarizationComparison}, right). A beneficial side-effect of this randomization is that it counteracts over-fitting to the training set, since the training images appear in many different forms, effectively creating a much larger dataset. In this sense, stochastic binarization is similar to dropout regularization \parencite{Hinton2012}. To capitalize on these benefits it is essential to redraw from the training data at the beginning of every epoch. Similarly, multiple draws from the validation and test sets should be used for model selection and evaluation in order to obtain robust results. 

\subsection{Model specifications}
\label{sec:ModelSpecifications}
We will evaluate HMCVI on the MNIST dataset with stochastic binarization for better comparability. The training data was resampled as described above before each iteration. For the validation and test set five random draws from the unbinarized sets were used. The HMCVI algorithm was implemented in python using the package Theano \parencite{Bergstra2010, Bastien2012}. All models were trained for several thousand epochs using Adam \parencite{Kingma2015} integrated with Theano by the package climin \parencite{Bayer2015}. Adam was run with the default parameters except for the step size, which was set to $10^{-4}$ or $5 \cdot 10^{-5}$.

In all experiments the decoding model $p(x_i|z_i)$ consisted of a conditionally independent Bernoulli distribution over the pixels with the rates given by a fully connected neural network with the latent variables as input. This network had two hidden layers with 200 neurons each and softplus ($\log(1 + \exp(x))$) activations. In the output layer the element-wise sigmoid activation function was applied. Similarly, for the initial encoder model $q_0(z_i|x_i)$ a multivariate normal distribution with diagonal covariance was used, where the parameters were given by a second neural network taking the observed variables as inputs. Again, two hidden layers with 200 units each were used, here with rectified linear unit (ReLU, $\max(0, x)$) activations. In the output layer the parameters corresponding to the mean were left unchanged, while the variance parameters were passed through the exponential function. As prior distribution for the latent variables a centred isotropic Gaussian distribution was chosen. 

For all HMCVI experiments the leapfrog method was applied and the step size learnt (constrained to be positive). In experiments without partial momentum update ($\alpha = 0$ fixed) the reverse momentum model $r_V$ and final momentum model $r_\textrm{final}$ were joined into a single model as explained in section \ref{sec:SimplificationWithoutPartialMomentumUpdate}. This model was like the initial encoder model, but with the position and the time step as additional inputs. If partial momentum updates were included, the final momentum model was as in the previous case, but the reverse momentum model $r_V$ was a separate network with the updated momentum as an additional input and otherwise the same specifications as before (see section \ref{sec:TransitionDensitiesNoAcceptance}).

Where an acceptance step was included, either the converged chain approximation ("simple") derived in section~\ref{sec:TransDensitiesWithAcceptReverse} was used for the reverse acceptance probability $\mathbb{P}(A = 1|S_t = s_t, t, x)$ or a neural network was trained ("NN") for it. The output of this neural network, whose final layer was passed through the $\tanh$ function, was added to the converged chain approximation and then clipped to be in $[0, 1]$. The network took the current state, the time step and the observed variables as inputs and consisted of two hidden layers with 200 units each and ReLU activations.

For the canonical momentum distribution, which also specifies the kinetic energy, a zero mean multivariate normal distribution with diagonal covariance matrix was assumed throughout. For the diagonal entries three choices were compared: They were either set to 1 ("Identity") or learnt globally ("Global") or specified by a neural network ("NN"), taking the observed variables as input. In the second case the exponential function was applied to unconstrained parameters to ensure positivity. The neural network in the third case had a single hidden layer with 200 units and a ReLU activation and the exponential function as output transfer.

All parameters were independently initialized from a Gaussian distribution $N(0, 0.01)$. In HMCVI experiments the generative model and initial encoder model were then copied from a previously trained VAE (the same for all HMCVI experiments with the same number of latent variables). With this initialization the HMCVI methods showed much better training results than with fully random initialization.

\subsection{Model comparison}

\begin{table*}[ht]
\centering

\begin{tabular}{lrrrrrrrr}
\toprule
Name & $d$ & \#HMC & \#LF & Partial & $M$ & Accept & $-\log(p(x)) \leq$ & $- \log(p(x)) \approx$ \tn 
\midrule
Basic VI 2D & 2 & 0 & 0 & - & - & - & 131.76 & 128.95 \tn 
HMCVI 1 & 2 & 1 & 4 & - & Global & - & 130.12 & 127.50 \tn 
HMCVI 2 & 2 & 1 & 12 & - & Global & - & 130.11 & 127.54 \tn 
HMCVI 3 & 2 & 2 & 6 & - & Global & - & 129.78 & 127.27 \tn 
HMCVI 4 & 2 & 3 & 4 & - & Global & - & 129.62 & 127.14 \tn 
HMCVI 5 & 2 & 3 & 4 & Yes & Global & - & 129.25 & 127.03 \tn 
HMCVI 6 & 2 & 3 & 4 & - & Identity & - & 129.59 & 127.11 \tn 
HMCVI 7 & 2 & 3 & 4 & - & NN & - & 129.32 & 127.06 \tn 
HMCVI 8 & 2 & 3 & 4 & Yes & NN & - & 128.96 & 126.94 \tn 
HMCVI 9 & 2 & 3 & 4 & - & Global & Simple & 129.93 & 127.24 \tn 
HMCVI 10 & 2 & 3 & 4 & - & Global & NN & 129.88 & 127.17 \tn 
\midrule
Basic VI 20D & 20 & 0 & 0 & - & - & - & 92.35 & 88.27 \tn 
HMCVI 11 & 20 & 1 & 12 & - & Global & - & 89.77 & 87.77 \tn 
HMCVI 12 & 20 & 2 & 6 & - & Global & - & 89.83 & 87.53 \tn 
HMCVI 13 & 20 & 3 & 4 & - & Global & - & 90.24 & 87.56 \tn 
HMCVI 14 & 20 & 3 & 4 & Yes & Global & - & 90.15 & 87.49 \tn 
HMCVI 15 & 20 & 3 & 4 & - & Identity & - & 91.08 & 87.65 \tn 
HMCVI 16 & 20 & 3 & 4 & - & NN & - & 90.23 & 87.30 \tn 
HMCVI 17 & 20 & 3 & 4 & Yes & NN & - & 89.72 & 87.44 \tn 
HMCVI 18 & 20 & 3 & 4 & - & Global & Simple & 91.40 & 87.28 \tn 
HMCVI 19 & 20 & 3 & 4 & - & Global & NN & 91.37 & 87.32 \tn 
HMCVI 20 & 20 & 3 & 4 & - & NN & Simple & 91.38 & 87.20 \tn 
\bottomrule
\end{tabular}

\caption{Comparison of the obtained lower bound and marginal log-likelihood estimates for different HMCVI configurations with a 2-dimensional (top) and a 20-dimensional latent space (bottom). \#HMC and \#LF give the number of used HMC and leapfrog steps respectively. The fifth column indicates, whether partial momentum updates were permitted. The sixth column gives the strategy used for the covariance matrix $M$ of the canonical momentum distribution and the seventh column, whether the acceptance step was included and, if so, what approach was used (as described in section \ref{sec:ModelSpecifications}). The last two columns report the lower bound $\mathcal{L_\textrm{aux}}$ and the estimated NLL on the test set.}
\label{tab:Results}
\end{table*}

We maximized the lower bound for various different setups of the HMCVI framework. Table~\ref{tab:Results} shows the results obtained with a two-dimensional latent space (see appendix~\ref{app:LatentVisualizations} for some visualizations) and with a 20-dimensional latent space. The NLL estimates given were obtained using importance sampling with 5000 samples (described in appendix~\ref{app:NLLestimateImportSampling}).

From comparing the results, obtained using only a parametric posterior approximation (Basic VI 2D and 20D), to the HMCVI results it is obvious, that any additional HMC steps greatly improve the estimation quality. 

For the two-dimensional latent space we see that increasing the length of the simulated trajectory improves the results and that resampling the momentum more frequently (i.e.\ performing more HMC steps) is also beneficial (compare HMCVI 1-4). From the nature of HMC both of these observations are to be expected, since longer trajectories allow further movement through the latent space and hence better exploration. Likewise, more HMC steps implies a longer Markov chain, which should thus be closer to convergence. A more intuitive explanation of the second observation is, that initially the simulated particles may have high potential energies and move down the potential energy landscape increasing their kinetic energy. If their large built-up kinetic energy is then reduced by the resampling of the momentum, they can not move out of the potential energy basin they have slid into. Conversely, if there is a less frequent resampling of the momentum, their built-up momentum may carry them out of the basin again on the other side, so that their potential energy has not decreased as much and correspondingly their joint likelihood $p(x, z)$ has not increased as much (compare figures~\ref{fig:HMC_MOTION_1hmc_12lf} and \ref{fig:HMC_MOTION_3hmc_04lf}).

Interestingly, for the 20-dimensional latent space the bound worsens in our experiments, when the momentum is resampled more frequently, while the estimated NLL improves (see HMCVI 11-13). So w.r.t.\ the real target, the NLL, more HMC steps are positive, but this is not reflected in the bound. An explanation for this phenomenon could be that the auxiliary reverse model is not flexible enough to capture the additional reverse densities (introduced by the addition of HMC steps) as tightly, leading to a poorer bound.

Allowing partial momentum updates and the covariance matrix to depend on the observed variables further improved the performance as expected (HMCVI 5, 7, 14 and 16). With a two-dimensional latent space their combination produced in the best performing model (HMCVI 8). For the 20-dimensional latent space, the combination (HMCVI 17) yielded the best bound, but not the best NLL estimate. Fixing the covariance matrix to be the identity (HMCVI 6 and 15) performed worse than learning it globally for the 20-dimensional case, but no different for the two-dimensional case. Understandably, with two dimensions a global rescaling is unlikely to change much.

For the two-dimensional latent space, including the acceptance step returned worse results, but as to be expected the more complicated reverse probability model (HMCVI 10) outperformed the approach, where the chain was assumed to have already converged (HMCVI 9). The weaker performance of HMCVI with acceptance step in this case is probably due to the fact, that the short chains being used here have not nearly converged to their invariant distribution yet. Therefore, the reduced mixing due to the rejection of proposals outweighs possible gains from the improved posterior approximation, since only with the acceptance step the chain will actually converge to the true posterior. 

A different picture, however, presents itself for the 20-dimensional latent space: Again the lower bound is worse, when the acceptance step is included (HMCVI 18 and 19), but regarding the NLL estimate the models learnt with the acceptance step outperform all other models. This means that the inclusion of the acceptance step improved the quality of the VAE. This indicates, that in the larger latent space it is beneficial to reject some proposed transitions in order to obtain a better approximation of the posterior and this improved approximation allows a better decoder to be learnt. The poor quality of the bound is presumably due to the lacking flexibility of the reverse model, which has to deal with more noise and more complicated distributions, if the acceptance step is included (see section~\ref{sec:TransDensitiesWithAcceptReverse}). By combining the acceptance step with the input-dependent kinetic energy (HMCVI 20) the learnt model could be further improved as expected.
\section{Conclusion and future work}
\label{sec:ConclAndFuture}
In this work we analysed the previously suggested integration of the HMC algorithm into VI, focussing in particular on its theoretical foundations. By exploiting the structure of the HMC algorithm; we were able to include the Metropolis-Hastings acceptance step in the algorithm, which was previously left out, without adding any new variables. Only including this acceptance step in the HMC algorithm ensures the convergence of the chain to the true posterior. In our experiments the lower bound obtained when the acceptance step was included, was worse than without the acceptance step. However, w.r.t.\ the negative log-likelihood the models with acceptance step were superior (for a realistically sized latent space). The improved approximation of the posterior due to the inclusion of the acceptance step thus leads to a better variational auto-encoder being learnt. By increasing the flexibility of the reverse model this should also become apparent in the variational lower bound.

For the simplified case without the acceptance step, a better performance was also achieved by allowing partial momentum updates in the HMC algorithm, a generalization of the algorithm reported to be particularly beneficial for shorter-than-optimal trajectories. Further, we utilized the possibility of learning continuous parameters of the HMC algorithm as part of the maximization of the lower bound to make these parameters input-dependent. In this way, the algorithm is automatically adjusted to the current input. This lead to better results in our experiments, both with and without the acceptance step. In this work we only allowed the mass matrix to depend on the observed variables, but other parameters, such as the step size, could also be made input-dependent, promising further improvements.

While the HMCVI algorithm improves the density estimation, it also requires significantly more computational effort than basic VI, in particular, if the acceptance step is included. Making the algorithm computationally more efficient, for example by propagating approximate distributions instead of sampling individual points, would remove this drawback and also allow for longer chains leading to better convergence.

Another interesting question regarding HMCVI is the role of the auxiliary reverse model. Its existence and flexibility are necessary ingredients to make the lower bound tight and the other models train properly, but really the learnt reverse model is not needed once training is completed. In this sense, valuable training time is used for something unwanted. Understanding the function of this model further may yield computational speed-ups or better density estimation by removing apparent restrictions resulting from the current reverse model specifications.

\subsection*{Acknowledgements}

This work has been supported in part by the TACMAN project, EC Grant agreement no.\ 610967, within the FP7 framework programme.

\printbibliography{}

\clearpage
\begin{appendices}

\section{Derivation of the reverse acceptance probability}
\label{app:DerivationOfReverseAcceptanceProbability}
If we let $A$ be the event of accepting the proposed transition in the previous HMC step, the probability $\mathbb{P}(A = 1|S_t = s_t, t, x)$ of accepting it given the current position can be related to the distribution of $S_{t-1}^*$ by considering
\begin{equation}
\begin{split}
&\mathbb{P}(A = 1|S_t = s_t, t, x) \\
&\quad= f_{A, S_t|T, X}(1, s_t| t, x)/f_{S_t|T, X}(s_t| t, x),
\end{split}
\end{equation}
where $f_{S_t|T, X}(s_t| t, x) = f_{A, S_t|T, X}(1, s_t| t, x) + f_{A, S_t|T, X}(0, s_t| t, x)$. These terms can then we rewritten using $p_\textrm{accept}(s)$ defined in equation~\eqref{eq:AcceptanceProbability}:
\begin{align}
\begin{split}
&f_{A, S_t|T, X}(1, s_t| t, x) \\
&\quad\qquad = f_{A, S_{t-1}^*|T, X}\big(1, revHD(s_t)| t, x\big) \\
&\quad\qquad = p_\textrm{accept}(revHD(s_t)) \\
&\quad\qquad\qquad \cdot f_{S^*_{t-1}|T, X}\big(revHD(s_t)| t, x\big)
\end{split} \\
\begin{split}
&f_{A, S_t|T, X}(0, s_t| t, x) \\
&\quad\qquad = f_{A, S_{t-1}^*|T, X}\big(0, (z_t, -v_t)| t, x\big) \\
&\quad\qquad = \big(1 - p_\textrm{accept}(z_t, -v_t)\big) \\
&\quad\qquad\qquad \cdot f_{S^*_{t-1}|T, X}\big((z_t, -v_t)| t, x\big)
\end{split}
\end{align}
Now, if $H(z_t, -v_t) \geq H(HD(z_t, -v_t))$ holds, $p_\textrm{accept}(z_t, -v_t) = 1$ and inserting this in the above gives that $\mathbb{P}(A = 1|S_t = s_t, t, x) = 1$. This means the move to $s_t$ must have been accepted.

If this is not the case, then the acceptance probability cannot be simplified further without reducing the flexibility of the model. In this case one would ideally learn an approximation for $\mathbb{P}(A=1|S_t = s_t, t, x)$, taking $s_t$, $x$ and the time point $t$ as inputs. A good starting point for this model can be obtained by assuming that the Markov chain has already converged. Under this assumption $S_{t-1}^*$ would follow the canonical distribution, so we would have $f_{S^*_{t-1}|T, X}(s| t, x) \propto \exp(-H(s))$. Inserting this in the above equations and noting, that $HD(z_t, -v_t) = revHD(z_t, v_t)$ due to the invertibility of HD and $H(z_t, -v_t) = H(z_t, v_t)$ due to the symmetry of the kinetic energy, yields
\begin{equation}
\begin{split}
\mathbb{P}(A = 1&|S_t = s_t, t, x) \\
&= \exp(-H(revHD(s_t)) + H(s_t))
\end{split}
\end{equation}

In a nutshell, if $H(revHD(s_t)) \leq H(s_t)$ holds, the previous move was always accepted. Otherwise, the probability needs to be learnt, but will tend towards $\exp(-H(revHD(s_t)) + H(s_t))$ as the chain converges.

\section{Likelihood estimation by importance sampling}
\label{app:NLLestimateImportSampling}

The marginal likelihood $p(x)$ is estimated using importance sampling by generating $S$ samples from some sampling distribution $p_\textrm{samp}(z|x)$ and using the following estimation:
\begin{equation}
\begin{split}
p(x) &= \E_{z \sim p_\textrm{samp}} \left[\frac{p(x|z) \cdot \pi(z)}{p_\textrm{samp}(z|x)} \right] \\
&\approx \frac{1}{S} \sum_{s=1}^S \frac{p(x|z_s) \cdot \pi(z_s)}{p_\textrm{samp}(z_s|x)} \textrm{ for } z_s \sim p_\textrm{samp}
\end{split}
\end{equation}
For this estimation to be efficient, it is important that the sampling distribution tightly covers the true posterior $p(z|x)$. To achieve this, the sampling distribution, chosen to be a multivariate Gaussian, was centred on an estimate of the mean of the true posterior, obtained by sampling five times from the HMC-enhanced posterior approximation. The covariance matrix was taken from the initial encoder $q_0(z|x)$. This returned low variance estimates of the marginal likelihood with little dependence on the number of samples $S$ for $S > 2000$.

\begin{figure}[hb]
\centering
\includegraphics[width=\columnwidth]{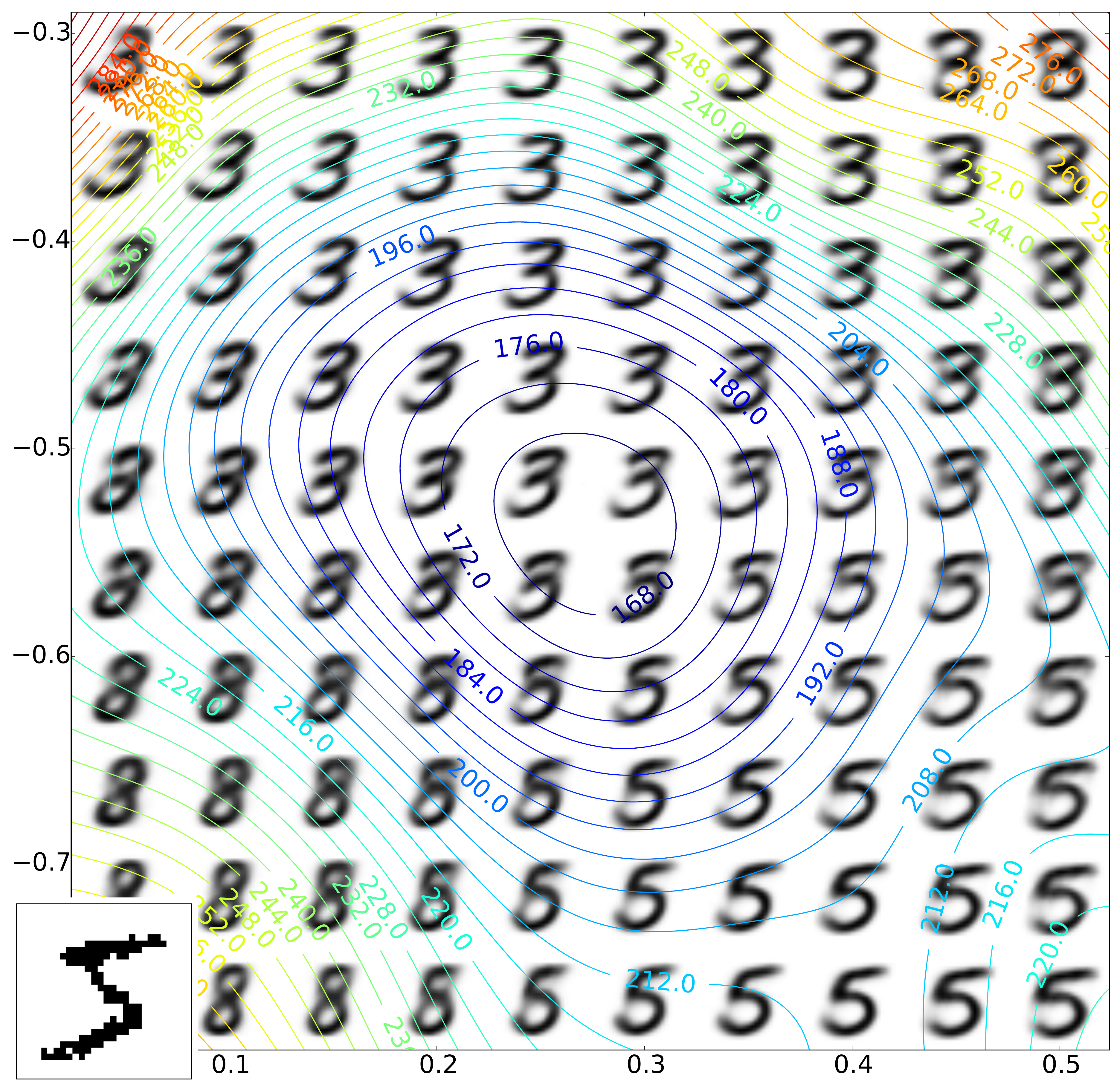}
\caption{Potential energy surface for the observed digit shown in the inset. The contours indicate the potential energy surface produced by a trained model with a 2-dimensional latent space. The plot also shows the mean images produced by the decoding model at evenly spaced points of the latent space.}
\label{fig:EnergySurfaceMNIST}
\end{figure}

\section{Visualizations of latent space}
\label{app:LatentVisualizations}

For each MNIST digit $x$ the potential energy surface given by $-\log p(x, z)$ differs. Figure~\ref{fig:EnergySurfaceMNIST} shows the energy surface produced by a trained model for a specific digit. For an intuitive understanding of the potential energy it also shows the mean images produced by the decoding model $p(x|z)$ at evenly spaced points in latent space. The closer the mean image is to the observed digit, the lower the potential energy.

\begin{figure}[hb]
\centering
\includegraphics[width=\columnwidth]{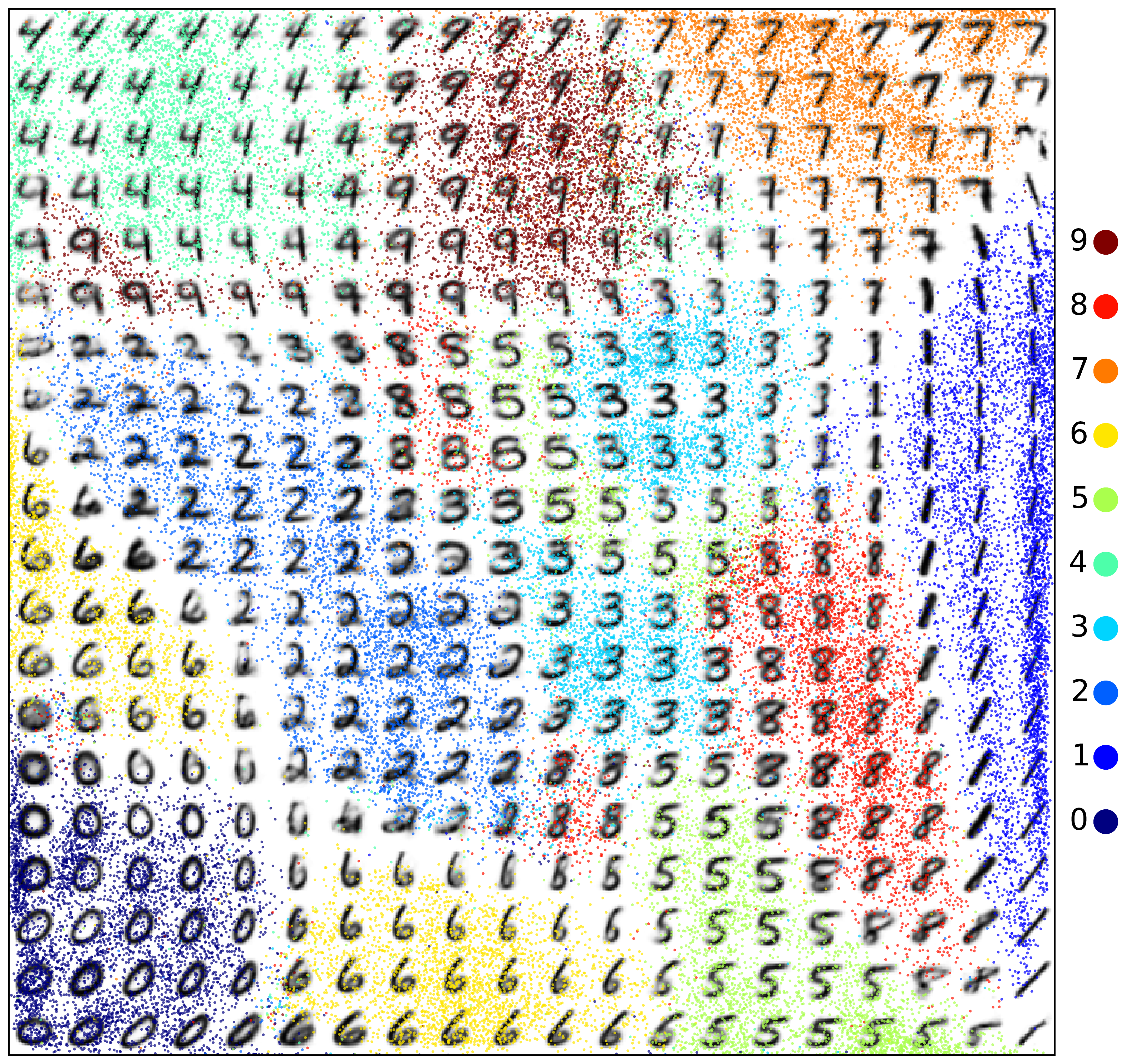}
\caption{Illustration of the two-dimensional latent space representation learnt by the model HMCVI 8 (see table~\ref{tab:Results}). To compensate for the Gaussian prior on the latent variables, linearly spaced coordinates in the unit square were transformed using the inverse Gaussian cdf. Therefore, the prior density in this view of latent space is uniform. For each coordinate the mean image produced by the decoder is shown. Additionally, the latent space representation of the training dataset as produced by the enhanced encoder is depicted (transformed by the Gaussian cdf), where each digit class is indicated by a different color.}
\label{fig:2d_latent_visualization}
\end{figure}

For the best performing model on two-dimensional latent space figure~\ref{fig:2d_latent_visualization} illustrates the learnt latent space, depicting both exemplary mean images produced by the decoding model $p(x|z)$ and the latent space coordinates of the training set under the learnt encoder (including the HMC steps). A clear (but not perfect) separation of the digits is immediately obvious, showing the power of this unsupervised model to capture structures in the data. Interestingly, the latent space is not occupied evenly, with transition areas between the digits completely vacant. With a more flexible decoder this behaviour should become less prominent.

\end{appendices}
\end{document}